\documentclass{article}

\usepackage[final]{neurips_2022}

\usepackage[utf8]{inputenc} % allow utf-8 input
\usepackage[T1]{fontenc}    % use 8-bit T1 fonts
\usepackage{hyperref}       % hyperlinks
\usepackage{url}            % simple URL typesetting
\usepackage{booktabs}       % professional-quality tables
\usepackage{amsfonts}       % blackboard math symbols
\usepackage{nicefrac}       % compact symbols for 1/2, etc.
\usepackage{microtype}      % microtypography
\usepackage[LGR,T1]{fontenc}

\usepackage[table]{xcolor}

\definecolor{tableHeader}{RGB}{55,126,184}
\definecolor{tableLineOne}{RGB}{245, 245, 245}
\definecolor{tableLineTwo}{RGB}{255, 255, 255}
\definecolor{specialgrey}{RGB}{90, 90, 90}

\usepackage{color}
\usepackage{algorithm}
\usepackage{algorithmic}
\usepackage{mathrsfs}
\usepackage{dsfont}
\usepackage{lmodern}
\usepackage{array}
\usepackage{bbm}
\usepackage{amsmath}
\usepackage{amssymb}
\usepackage[mathscr]{eucal}
\usepackage{graphicx}
\usepackage{mathrsfs}
\usepackage{psfrag}
\usepackage{color}
\usepackage{here}
\usepackage{wasysym}
\usepackage{enumitem}
\usepackage{subcaption}
\usepackage{amsthm}
\usepackage{lipsum}
\usepackage{todonotes}
\usepackage{tabulary}
\usepackage{pifont}
\usepackage{outlines}

\newcommand{\Exp}{\mathds{E}}

\newcommand{\Prob}{\mathds{P}}
\newcommand{\Real}{\mathds{R}}

\newcommand{\Ec}{\mathcal{E}}

\newcommand{\E}{\mathbb{E}}

\newcommand{\KL}{\mathbf{d}_{\mathrm{KL}}}

\newcommand{\data}{\mathcal{D}}

\newcommand{\reals}{\mathbf{R}}
\newcommand{\environment}{\mathcal{E}}

\DeclareMathOperator*{\softmax}{softmax}

\definecolor{ian_highlight}{RGB}{100, 2, 2}

\errorcontextlines\maxdimen

\makeatletter
    \newcommand*{\algrule}[1][\algorithmicindent]{\makebox[#1][l]{\hspace*{.5em}\thealgruleextra\vrule height \thealgruleheight depth \thealgruledepth}}%
\newcommand*{\thealgruleextra}{}
\newcommand*{\thealgruleheight}{.75\baselineskip}
\newcommand*{\thealgruledepth}{.25\baselineskip}

\newcount\ALG@printindent@tempcnta
\def\ALG@printindent{%
    \ifnum \theALG@nested>0% is there anything to print
        \ifx\ALG@text\ALG@x@notext% is this an end group without any text?
        \else
            \unskip
            \addvspace{-1pt}% FUDGE to make the rules line up
            \ALG@printindent@tempcnta=1
            \loop
                \algrule[\csname ALG@ind@\the\ALG@printindent@tempcnta\endcsname]%
                \advance \ALG@printindent@tempcnta 1
            \ifnum \ALG@printindent@tempcnta<\numexpr\theALG@nested+1\relax% can't do <=, so add one to RHS and use < instead
            \repeat
        \fi
    \fi
    }%
\usepackage{etoolbox}
\makeatother

\newbox\statebox
\newcommand{\myState}[1]{%
    \setbox\statebox=\vbox{#1}%
    \edef\thealgruleheight{\dimexpr \the\ht\statebox+1pt\relax}%
    \edef\thealgruledepth{\dimexpr \the\dp\statebox+1pt\relax}%
    \ifdim\thealgruleheight<.75\baselineskip
        \def\thealgruleheight{\dimexpr .75\baselineskip+1pt\relax}%
    \fi
    \ifdim\thealgruledepth<.25\baselineskip
        \def\thealgruledepth{\dimexpr .25\baselineskip+1pt\relax}%
    \fi
    \State #1%
    \def\thealgruleheight{\dimexpr .75\baselineskip+1pt\relax}%
    \def\thealgruledepth{\dimexpr .25\baselineskip+1pt\relax}%
}
\newcount\Comments
\newcommand{\kibitz}[2]{\ifnum\Comments=1{\textcolor{#1}{\textsf{\footnotesize #2}}}\fi}

\newcommand{\githubtestbed}{\url{https://github.com/deepmind/neural\_testbed}}

\title{The Neural Testbed:\\ Evaluating Joint Predictions}

\author{%
  Ian Osband\thanks{Contact \texttt{iosband@deepmind.com}},
  Zheng Wen,
  Seyed Mohammad Asghari, \\
  \textbf{
  Brendan O'Donoghue,
  Botao Hao,
  Dieterich Lawson,
  Morteza Ibrahimi,} \\
  \textbf{
  Xiuyuan Lu,
  Vikranth Dwaracherla,
  and Benjamin Van Roy} \\
  DeepMind, Efficient Agent Team, Mountain View
}

\begin{document}

\maketitle

\begin{abstract}
Predictive distributions quantify uncertainties ignored by point estimates.
This paper introduces \textit{The Neural Testbed}: an open-source benchmark for controlled and principled evaluation of agents that generate such predictions.
Crucially, the testbed assesses agents not only on the quality of their marginal predictions per input, but also on their joint predictions across many inputs.
We evaluate a range of agents using a simple neural network data generating process.
Our results indicate that some popular Bayesian deep learning agents do not fare well with joint predictions, even when they can produce accurate marginal predictions.
We also show that the quality of joint predictions drives performance in downstream decision tasks.
We find these results are robust across choice a wide range of generative models, and highlight the practical importance of joint predictions to the community.
\end{abstract}

%%%%%%%%%%%%%%%%%%%%%%%%%%%%%%%%%%%%%%%%%%%%%%%%%%%%%%%%%%%%%%%%%%%%%%%%%%%%%%%%%%%%%%%%%%% INTRODUCTION
%%%%%%%%%%%%%%%%%%%%%%%%%%%%%%%%%%%%%%%%%%%%%%%%%%%%%%%%%%%%%%%%%%%%%%%%%%%%%%%%%%%%%%%%%%%
\section{Introduction}
\label{sec:intro}

% If you want to do well in decisions, you need to have good joint predictions.
Most work on supervised learning has focused on marginal predictions.
Marginal predictions predict one label given one input, but do not model the dependence between multiple predictions.
For decision making, it is not enough to have good marginal predictions; the quality of \textit{joint} predictions drives decision performance \citep{wen2022predictions}.
Joint predictions predict multiple labels given multiple inputs, and may capture some correlation between outcomes.
This distinction can be particularly important in learning settings where joint predictions allow an agent to distinguish what it knows from what it does not know \citep{li2011knows,lu2021reinforcement}.

\begin{figure}[!h]
  \centering
  \includegraphics[width=0.75\linewidth]{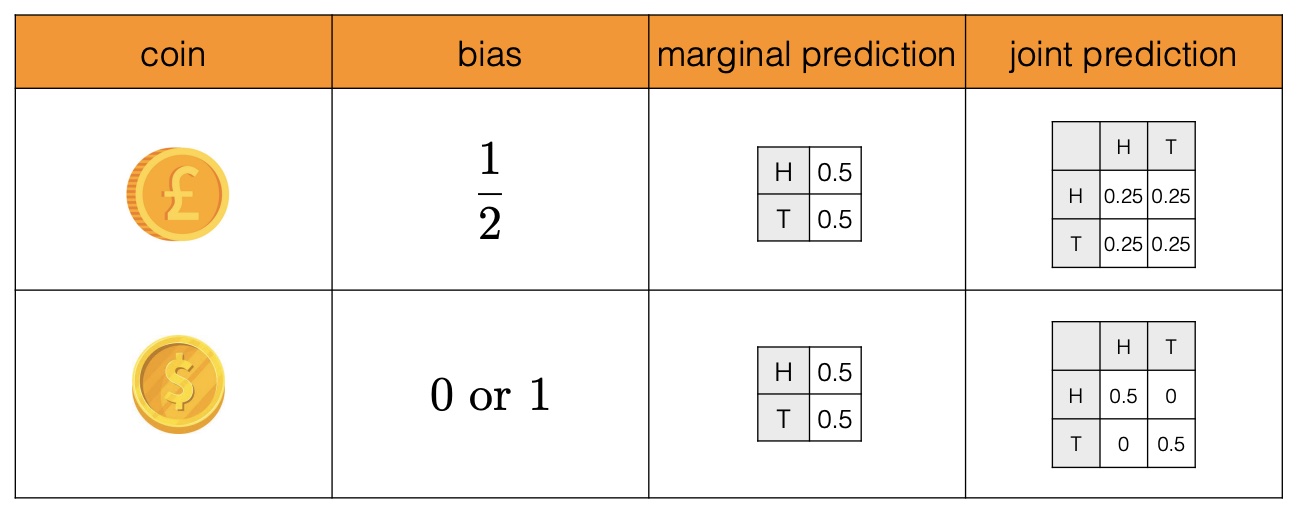}
  \caption{Two coins with identical marginal predictions, but distinguished by joint predictions.}
\label{fig:coin_example}
\end{figure}

% Coin example to explain the difference.
Figure~\ref{fig:coin_example} presents a stylized example designed to highlight the importance of joint predictions in decision making.
Consider two coins `£' and `\$' with different \textit{bias}=`probability of heads'.
Coin £ has a known bias of $\frac{1}{2}$, whereas coin \$ has an unknown bias of either 0 or 1, and which are both equally likely.
Examining the marginal prediction over a single flip: the two coins present identical outcomes 50:50.
However, if we consider the outcome over two successive flips, which can be modeled as a two-by-two grid, then the difference between these coins is evident in their joint predictions.
If you want to maximize the cumulative heads through time, then it's important to know the difference between these two settings.
In this case, a learning agent should first choose \$ and then, depending on the outcome of that flip heads/tails, employ a fixed policy of \$/£ forward.
Marginal predictions alone cannot drive this sort of policy, since they do not distinguish the two coins \citep{wen2022predictions}.

% Motivated by grand challenges, but it's important to have simple examples to benchmark and understand.
Our research is motivated by the grand challenges in artificial intelligence, and the great progress that has been made in deep learning systems \citep{krizhevsky2012imagenet,brown2020language}.
However, as these systems move beyond prediction and towards actually making decisions we have very little understanding of how and where popular deep learning approaches are suitable for joint predictions and hence decision making \citep{mnih15nature, silver2016alphago}.
To this end, we introduce \textit{The Neural Testbed} as a simple and clear benchmark for evaluating the quality of joint predictions in deep learning systems.
This work is meant to be a `sanity check' for popular deep learning approaches in a simple setting, and one that can help guide future research.

% Neural Testbed is the problem of random MLP, the gold standard is Bayesian.
% But you don't need to be Bayesian to use it.
% \fillpara
The Neural Testbed works by generating random classification problems using a neural-network-based generative process.
The testbed splits data into a training set and testing set, allows a deep learning agent to train on the training set, and then evaluates the quality of the predictions on the testing set.
It is worth noting that the problem framed by the Testbed is a {\it computational} one.
Optimal performance would be attained by carrying out exact Bayesian inference: given infinite compute time, an agent could calculate the posterior distribution, which maximizes performance.
However, due to the complexity of the data generating process, this is infeasible.
The agents we study serve as approximate inference algorithms, and we can compare their performance purely through the quality of their predictions, without worrying `is XYZ Bayesian?' \citep{izmailov2021bayesian}.

\begin{figure}[!h]
  \centering
  \includegraphics[width=0.9\linewidth]{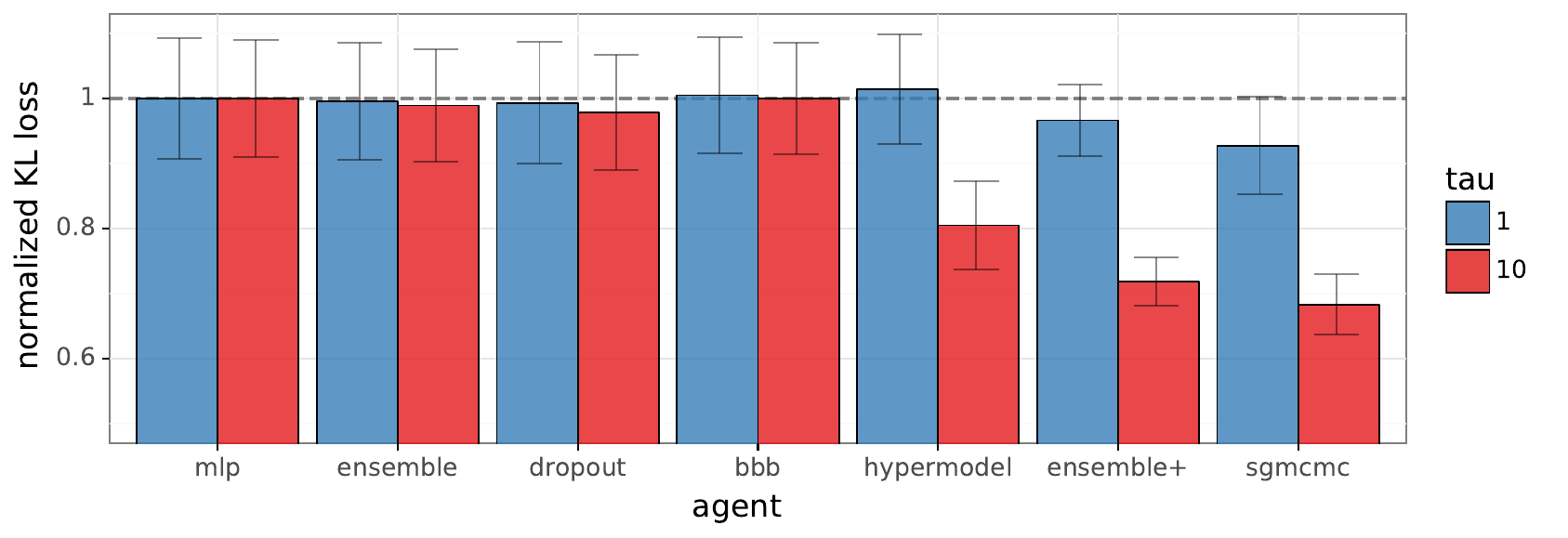}
  \caption{Quality of marginal and joint predictions on Neural Testbed (Section~\ref{sec:results_higher}).}
\label{fig:overall_performance_both}
\end{figure}

% This simple setting shows some really important results, joint versus marginal.
Figure~\ref{fig:overall_performance_both} offers a preview of our results in Section~\ref{sec:results}, where we compare benchmark approaches to Bayesian deep learning.
This plot shows the KL loss when making $\tau$ simultaneous predictions.
We compare the quality of marginal ($\tau=1$) and joint ($\tau=10$) predictions, normalized so that and MLP has loss=1.
We see that, after tuning, most Bayesian deep learning approaches do not significantly outperform a single MLP in marginal predictions.

However, once we examine joint predictive distributions of order $\tau=10$, there is a clear difference in performance among benchmark agents.
In particular, some of the most popular benchmark approaches to Bayesian deep learning (\texttt{ensemble} \citep{lakshminarayanan2017simple}, \texttt{dropout} \citep{Gal2016Dropout}, \texttt{bbb} \citep{blundell2015weight}) do not outperform the baseline MLP when evaluated in joint predictions.
At the same time, there are other approaches that perform much better in terms of joint predictions in this simple synthetic challenge.
We will go on to show that these same agents perform better in decision making, and that these observations are robust to choice of generative model.

%%%%%%%%%%%%%%%%%%%%%%%%%%%%%%%%%%%%%%%%%%%%%%%%%%%%%%%%%%%%%%%%%%%%%%%%%%%%%% Key contributions
\vspace{-3mm}
\subsection{Key contributions}
\label{sec:key-contributions}
% \vspace{-2mm}

% Frame the problem of joint predictions in random MLP.
% Work helps to bridge theory and practice.
% \fillpara
\textbf{We introduce \textit{The Neural Testbed}, a simple benchmark for the field that involves making predictions in a neural-network-based generative model.}
This work helps to bridge theory and practice, and provide an objective metric to assess the quality of approximate posterior inference in neural networks.
We are the first paper to propose a concrete evaluation procedure for the quality of joint predictions in neural network classification.
% Highly optimized opensource code, reference agents, accessible.
% \fillpara

\textbf{Together with this conceptual contribution, we open-source code in Appendix~\ref{app:code}}.
This consists of highly optimized evaluation code, reference agent implementations and automated reproducible analysis.
The testbed uses JAX internally \citep{jax2018github}, but can be used to evaluate any python agent.
We believe that this library will be a major contribution to researchers and, due to its low computational cost, a boon to accessibility.

% Big results that BDL approaches are bad.
% This was meant to be the simplest possible sanity check, but it doesn't even work here.
% Should be an alarm bell to the community.
% \fillpara
We use this new benchmark to obtain some important new experimental results.
\textbf{We discover that several of the most popular approaches to Bayesian deep learning do not perform well at joint prediction}, and highlight this issue to the community.
Further, we show that there \textit{are} alternative approaches that \textit{do} perform well in terms of joint prediction.
% These results carry over to decision making and bandit.
Prior work has suggested that, in theory, the quality of joint predictions can drive decision performance \citep{wen2022predictions}.
In this paper we provide empirical evidence that this effect occurs in practical deep learning systems.
\textbf{We observe that performance in a neural bandit is highly correlated with performance in joint prediction}, and that it is not significantly correlated with the quality of marginal predictions.

% Results are robust across choice of generative model.
Finally, \textbf{we show that the results in this paper are robust to the variations in the data generating model}.
Although we focus on a 2-layer ReLU MLP with 50 hidden units for most of our experiments, the results we obtain are highly correlated across a wide range of alternative activation functions or network widths.
This robustness supports the view that the field should be aware of these issues in joint prediction, and may help to stimulate future research in this area.
Follow-up work has gone on to show that these results also carry over to challenge datasets popular in the community \citep{osband2022evaluating}.

%%%%%%%%%%%%%%%%%%%%%%%%%%%%%%%%%%%%%%%%%%%%%%%%%%%%%%%%%%%%%%%%%%%%%%%%%%%%%% Related work
\vspace{-1mm}
\subsection{Related work}
\label{sec:intro_related}
\vspace{-2mm}

% \1 A lot of people have been looking into uncertainty in deep learning... but there's uncertainty in what works.
%     \2 List some of the agents papers
%     \2 Some of the dropout-scandal papers.
There is a rich literature around uncertainty estimation in deep learning.
Much of this work has focused on agent development, with a wide variety of approaches including variational inference \citep{blundell2015weight}, dropout \citep{Gal2016Dropout}, ensembles \citep{osband2015bootstrapped, lakshminarayanan2017simple}, and MCMC \citep{welling2011bayesian, hoffman2014no}.
However, even when approaches become popular within particular research communities, there are still significant disagreements over the quality of the resultant uncertainty estimates \citep{osband2016risk, hron2017variational}.

% \1 There are a lot of benchmarks out there, but most of these are looking at marginals.
%     \2 Classic datasets
%     \2 Bayesian deep learning competitions etc
Bayesian deep learning has largely relied on benchmark problems to guide agent development and measure agent progress.
These typically include classic deep learning datasets but supplement the usual goal of classification accuracy to include an evaluation of the probablistic predictions via negative log likelihood (NLL) and expected calibration error (ECE) \citep{nado2021uncertainty}.
More recently, several efforts have been made to supplement these datasets with challenges tailored towards Bayesian deep learning, and explicit Bayesian inference \citep{wilson2021evaluating}.
This literature has largely focused on evaluating marginal predictions, paired with evaluation on downstream tasks \citep{riquelme2018deep}.
Our work is motivated by the importance of \textit{joint} predictions in driving good performance in sequential decisions \citep{wen2022predictions}.
We share motivation with the work of \citet{wang2021beyond}, but show that directly measuring joint likelihoods can provide new information beyond marginals.
Follow up work has built upon the research in our paper, to extend the analysis of joint distributions to higher-order joint distributions, and empirical datasets \citep{osband2022evaluating}.

%%%%%%%%%%%%%%%%%%%%%%%%%%%%%%%%%%%%%%%%%%%%%%%%%%%%%%%%%%%%%%%%%%%%%%%%%%%%%%%%%%%%%%%%%%% PREDICTIVE DISTRIBUTIONS
%%%%%%%%%%%%%%%%%%%%%%%%%%%%%%%%%%%%%%%%%%%%%%%%%%%%%%%%%%%%%%%%%%%%%%%%%%%%%%%%%%%%%%%%%%%
\vspace{-1mm}
\section{Evaluating predictive distributions}
\label{sec:beyond_marginals}
\vspace{-2mm}
% \begin{outline}
% \1 This is a paragraph that is going to summarize what happens in this section.
%     \2 Describe the problem formulation.
%     \2 Describe computational tools.
%     \2 Talk about why this is so important.
% \end{outline}  
% In this section, we introduce notation for the standard supervised learning framework we will consider, which involves classification, as well as our evaluation metric.  Our evaluation metric is equivalent to cross entropy loss, except that we will focus on the difference between that and the minimum possible cross entropy loss.  This difference is the Kullback–Leibler loss (KL-loss), which is more natural to interpret and compare across environments.  Where we differ substantially from the supervised learning literature is in evaluating joint predictive distributions, not just marginals, based on this loss.  We also explain in this section how we approximate KL-loss for high-order predictions using random partitions and Monte Carlo simulation.

In this section, we introduce notation for the standard supervised learning we consider as well as our evaluation metric: KL-loss.
We review the distinction between marginal and joint predictions, and numerical schemes to estimate KL divergence via Monte Carlo sampling.
% We also explain how we estimate the KL-loss for high-order joint predictions where exact computation is infeasible.

\begin{minipage}{.5\linewidth}
{\small
    \begin{algorithm}[H]
    \caption{KL-Loss Estimation}
    \label{alg:kl-computation}
    \begin{algorithmic}
    \FOR{$j=1,2,\ldots,J$}
    \STATE sample environment and training data
    \STATE train agent on training data
    \FOR{$n=1,2,\ldots,N$}
    \STATE sample $\tau$ test data pairs
    \STATE compute environmentlikelihood $p_{j,n}$
    \STATE compute agent likelihood $\hat{p}_{j,n}$
    \ENDFOR
    \ENDFOR
    \RETURN $\frac{1}{JN} \sum_{j=1}^J \sum_{n=1}^N \log \left(p_{j,n} / \hat{p}_{j,n} \right)$
    \end{algorithmic}
    \end{algorithm}
}
\end{minipage}
\hfill
\begin{minipage}{.45\linewidth}
    \begin{figure}[H]
        \centering
          \includegraphics[width=.95\linewidth]{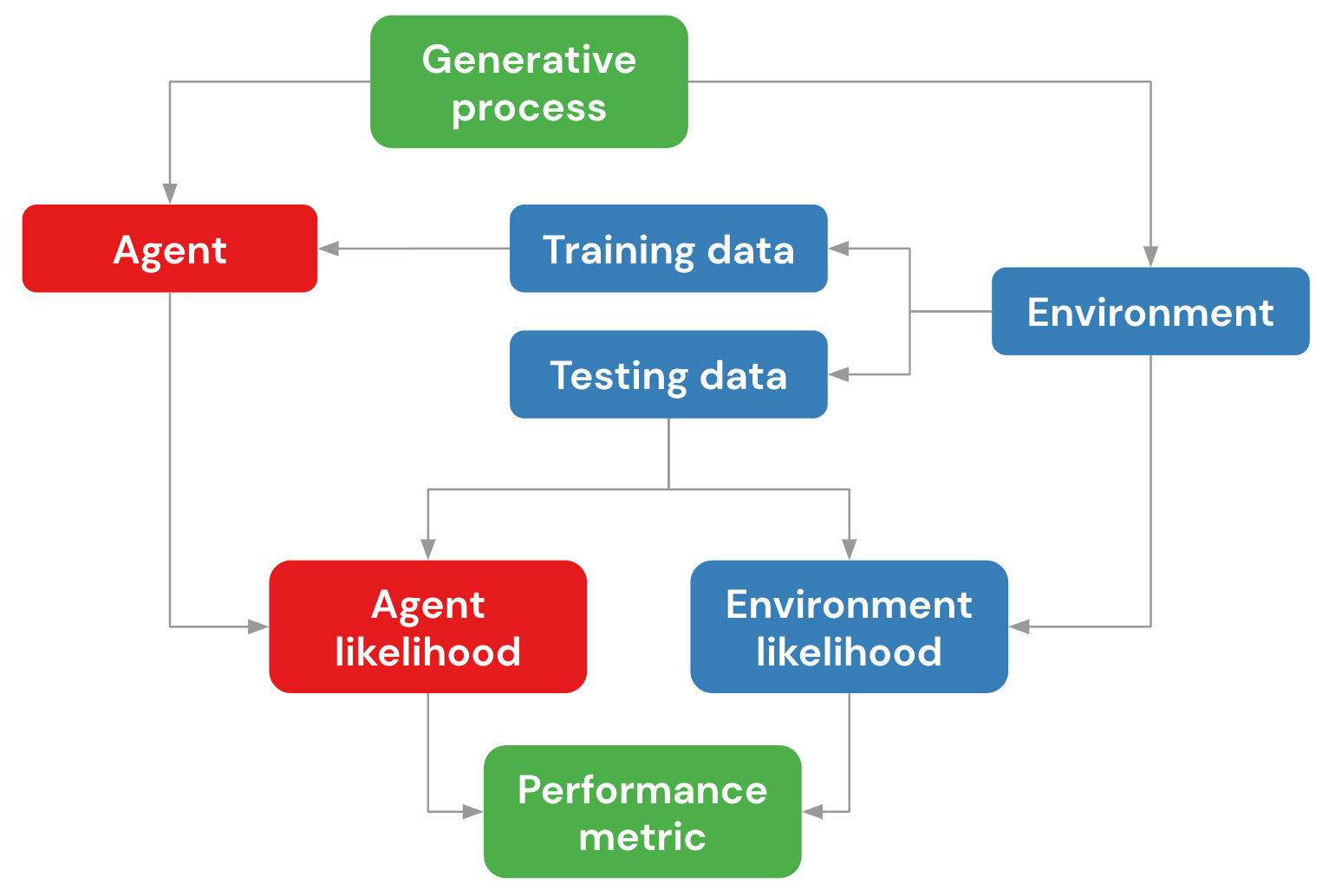}
        %   \vspace{-2mm}
          \caption{The Neural Testbed}
        \label{fig:testbed-diaigram}
    \end{figure}
\end{minipage}

%%%%%%%%%%%%%%%%%%%%%%%%%%%%%%%%%%%%%%%%%%%%%%%%%%%%%%%%%%%%%%%%%%%%%%%%%%%%%% Formulation
\subsection{Environment and predictions}
\label{sec:data-generation}
Consider a sequence of pairs $((X_t, Y_{t+1}): t =0,1,2,\ldots)$, where each $X_t$ is a feature vector and each $Y_{t+1}$ is its target label.  Each target label $Y_{t+1}$ is produced by an {\it environment} $\environment$, which we formally take to be a conditional distribution $\environment(\cdot|X_t)$.  The environment $\environment$ is a random variable; this reflects the agent's uncertainty about how labels are generated.  Note that $\Prob(Y_{t+1} \in \cdot | \environment, X_t) = \environment(\cdot|X_t)$ and $\Prob(Y_{t+1} \in \cdot | X_t) = \E[\environment(\cdot|X_t) | X_t]$.

We consider an agent that learns about the environment from training data \mbox{$\data_T \equiv ((X_t, Y_{t+1}): t =0,1,\ldots, T-1)$}.
After training, the agent predicts testing class labels $Y_{T+1:T+\tau} \equiv (Y_{T+1}, \dots, Y_{T+\tau})$ from unlabeled feature vectors \mbox{$X_{T:T+\tau-1} \equiv (X_T, \dots, X_{T+\tau-1})$}.

We describe the agent's predictions in terms of a generative model, parameterized by a vector $\theta_T$ that the agent learns from the training data $\data_T$.  For any inputs $X_{T:T+\tau-1}$, $\theta_T$ determines a predictive distribution, which could be used to sample imagined outcomes $\hat{Y}_{T+1:T+\tau}$.  Hence, the agents $\tau^{\rm th}$-order predictive distribution is given by
$$\hat{P}_{T+1:T+\tau} = \Prob(\hat{Y}_{T+1:T+\tau} \in \cdot | \theta_T, X_{T:T+\tau-1}),$$
which represents an approximation to what would be obtained by conditioning on the environment:
$$P^*_{T+1:T+\tau} = \Prob \left(Y_{T+1:T+\tau} \in \cdot \middle | \environment , X_{T:T+\tau-1} \right).$$
If $\tau=1$, this represents a marginal prediction; that is a prediction of a label for a single input.
For $\tau>1$, this is a joint prediction over labels for $\tau$ different inputs.

%%%%%%%%%%%%%%%%%%%%%%%%%%%%%%%%%%%%%%%%%%%%%%%%%%%%%%%%%%%%%%%%%%%%%%%%%%%%%% KL divergence
\subsection{Kullback–Leibler loss}
\label{sec:KL-loss}

We use expected KL-loss to quantify the error between an agent's predictive distribution $\hat{P}_{T+1:T+\tau}$ and the prescient prediction $P^*_{T+1:T+\tau}$ that would be made given full knowledge of the environment:
\begin{align}
\label{eq:kl_tau}
\KL^\tau =& \, \E \big[ 
    \KL \big(P^*_{T+1:T+\tau}
    \big \|
    \hat{P}_{T+1:T+\tau}
    \big)
    \big].
\end{align}
The expectation is taken over all random variables, including the environment $\environment$, the parameters $\theta_T$, $X_{T:T+\tau-1}$, and $Y_{T+1:T+\tau}$.
Note that $\KL^\tau$ is equivalent to the widely used notion of cross-entropy loss, though offset by a quantity that is independent of $\theta_T$.

In contexts we will consider, it is not possible to compute $\KL^\tau$ exactly.
As such, we will approximate $\KL^\tau$ via Monte Carlo simulation, as described by Algorithm \ref{alg:kl-computation}.  First, a set of environments is sampled.  Then, for each sampled environment, a training dataset is sampled.  For sampled environment and corresponding training data set, the agent is re-initialized, trained, and then tested on $N$ independent test data $\tau$-samples.  Note that each test data $\tau$-sample includes $\tau$ data pairs.  For each test data $\tau$-sample, the likelihood of the environment $p_{j,n}$ is computed exactly, but that of the agent's predictive distribution is approximated via another Monte Carlo simulation, and we use $\hat{p}_{j,n}$ to denote this approximation.  The estimate of  $\KL^\tau$ is taken to be the sample mean of these  log-likelihood ratios.

%%%%%%%%%%%%%%%%%%%%%%%%%%%%%%%%%%%%%%%%%%%%%%%%%%%%%%%%%%%%%%%%%%%%%%%%%%%%%%%%%%%%%%%%%%% NEURAL TESTBED
%%%%%%%%%%%%%%%%%%%%%%%%%%%%%%%%%%%%%%%%%%%%%%%%%%%%%%%%%%%%%%%%%%%%%%%%%%%%%%%%%%%%%%%%%%%
\vspace{-2mm}
\section{The Neural Testbed}
\label{sec:neural_testbed}
\vspace{-1mm}

% \begin{outline}
% \1 This is a section to outline the neural testbed.
%     \2 Why do we need a testbed?
%     \2 Describe the opensource code at high level.
%     \2 Describe the generative model.
%     \2 Describe the benchmark agents.
% \end{outline}

In this section we introduce the Neural Testbed.
We believe that a simple, clear and accessible testbed can provide significant value to community.
We provide a high-level overview of the open-source code which we release in Appendix~\ref{app:code}.
We then provide more details on the underlying generative model, together with an extensive selection of benchmark agents that we have tuned to perform well in this setting.

% \begin{figure}[!ht]
%   \centering
%   \includegraphics[width=.95\linewidth]{figures/neural_testbed_diagram5.png}
%   \vspace{-2mm}
%   \caption{Overview of the Neural Testbed. For each random seed the testbed samples an environment realization with training and testing data. The testbed then compares the estimated log-likelihood of the true environment and that of a trained agent to estimate KL divergence.}
% \label{fig:testbed-diaigram}
% \vspace{-4mm}
% \end{figure}

%%%%%%%%%%%%%%%%%%%%%%%%%%%%%%%%%%%%%%%%%%%%%%%%%%%%%%%%%%%%%%%%%%%%%%%%%%%%%% Generative model
\vspace{-1mm}
\subsection{Synthetic data generating processes}
\label{sec:generative}
\vspace{-1mm}

By data generating process, we do not mean only the conditional distribution of data pairs $(X_t,Y_{t+1})|\environment$ but also the distribution of the environment $\environment$.  
%Hence, a data generating process specifies the joint distribution of $(\environment, X_t, Y_{t+1})$. Recall that we assume $(X_t:t=0,1,\ldots)$ to be i.i.d. and $((X_t,Y_{t+1}):t=0,1,\ldots)$ to be i.i.d. conditioned on $\environment$; these assumptions hold for the data generating processes of the Testbed.  
% The Testbed considers 2-dimensional inputs and binary classification problems, although the generating processes can be easily extended to any input dimension and number of classes.
% The Testbed offers three data generating processes distinguished by a ``temperature'' setting, which signifies the signal-to-noise ratio (SNR) of the generated data. The agent can be tuned separately for each setting.  This reflects prior knowledge of whether the agent is operating in a high SNR regime such as image recognition or a low SNR regime such as weather forecasting.
The Testbed considers 2-dimensional inputs and binary classification problems.
The logits are sampled from a 2-hidden-layer ReLU MLP with (50,50) hidden units and Xavier initialization \citep{glorot2010understanding}.
We choose this process to be maximally simple and canonical in the deep learning world.
However, we will go on to show that the key findings of this paper are not particularly sensitive to the exact choice of generative model.

% To generate a model, the Testbed samples a 2-hidden-layer ReLU MLP with 2 output units, which are scaled by $1/\rm{temperature}$ and passed through a softmax function to produce class probabilities. The MLP is sampled according to standard Xavier initialization \citep{glorot2010understanding}, with the exception that biases in the first layer are drawn from $N(0, \frac{1}{\text{input dim}})$. The inputs $(X_t:t=0,1,\ldots)$ are drawn i.i.d. from $N(0, I)$. The agent is provided with the data generating process as prior knowledge. 

% In Section \ref{sec:KL-loss}, we described KL-loss as a metric for evaluating performance of an agent. %that engages with a fixed data generating process and trains on a fixed number of data pairs. In particular, the expectation in Equation \ref{eq:kl_tau} is with respect to a probability distribution that depends on the data generating process. 
The Neural Testbed estimates KL-loss, with $\tau  \in \{1,10\}$, for three temperature settings and several training dataset sizes.
The temperature $\rho$ controls the signal to noise ratio as the class probabilities are given by $\softmax({\text{logits}} / \rho)$.
%For each combination of $(\rm{temperature}, \rm{train\_size}, \tau)$, the Testbed sweeps over $10$ random seeds, generating independent MLP and data for each. 
For each value of $\tau$, the KL-losses are averaged to produce an aggregate performance measure. 
Further details concerning data generation and agent evaluation are offered in Appendix \ref{app:testbed_pseudo_code}.

%%%%%%%%%%%%%%%%%%%%%%%%%%%%%%%%%%%%%%%%%%%%%%%%%%%%%%%%%%%%%%%%%%%%%%%%%%%%%% Why synthetic data?
\vspace{-1mm}
\subsection{Why do we need a synthetic testbed?}
\label{sec:why_synthetic}
\vspace{-1mm}

% \1 Talking about simplicity and clarity.
The Neural Testbed is designed to be a maximally simple problem that investigates the key properties of uncertainty modeling in deep learning.
Progress in deep learning has been driven both by challenge datasets that stretch agent capabilities \citep{deng2009imagenet, krizhevsky2012imagenet}, together with foundational work that builds understanding \citep{bartlett2021deep}.
In this work, we provide a benchmark designed to improve our \textit{understanding} of probabilistic predictions beyond marginals.
Doing well in the testbed is not necessarily an impressive grand success in AI, although doing poorly in such a simple setting may reveal fundamental flaws in algorithm design.

% \1 Avoiding overfitting to fixed datasets.
A key property of the testbed is that it is specified by a probabilistic model, rather than a finite collection of datasets.
Benchmarks that rank performance on datasets are vulnerable to overfitting through iterative hill-climbing on the data included in the benchmark \citep{russo2016controlling}, which may not generalize to data outside of the benchmark \citep{recht2018cifar}.
In contrast, access to a generative model means that we can produce an unlimited amount of testing data from our problem of interest.
We can avoid the dangers of overfitting to any specific choices of benchmark dataset simply by generating more samples.

%%%%%%%%%%%%%%%%%%%%%%%%%%%%%%%%%%%%%%%%%%%%%%%%%%%%%%%%%%%%%%%%%%%%%%%%%%%%%% Benchmark agents
\vspace{-1mm}
\subsection{Benchmark agents}
\label{sec:agents}
\vspace{-1mm}

Table~\ref{tab:agent_summary} lists agents that we study and compare as well as hyperparameters that we tune.  In our experiments, we optimize these hyperparameters via grid search.
Our implementations, which aim to match `canonical' versions, are available in Appendix~\ref{app:code}. 
% Further detail on these agents is provided in Appendix \ref{app:agents}.

% \bgroup
% \rowcolors{2}{tableLineOne}{white}

In addition to these agent implementations, our open-source offerings include all the evaluation code to reproduce the results of this paper.
Our experiments make extensive use of parallel computation to facilitate hyperparameter sweeps.
Nevertheless, the overall computational cost is relatively low by modern deep learning standards and relies only on standard CPUs.
For reference, evaluating the \texttt{mlp} agent across all the problems in our testbed requires less than 3 CPU-hours.
We view our open-source effort as a substantial contribution of this work.

\begin{table}[!th]
\vspace{-3mm}
\caption{Summary of benchmark agents, full details in Appendix \ref{app:agents}.}
\begin{center}
\label{tab:agent_summary}
\resizebox{\columnwidth}{!}{%
\begin{tabular}{|l|l|l|}
\hline
\rowcolor{tableHeader}
\textcolor{white}{\textbf{agent}}          & \textcolor{white}{\textbf{description}}            & \textcolor{white}{\textbf{hyperparameters}} \\[0.5ex]  \hline
\textbf{\texttt{mlp}}            & Vanilla MLP        &  $L_2$ decay                        \\
\textbf{\texttt{ensemble}}       & `Deep Ensemble' \citep{lakshminarayanan2017simple}          & $L_2$ decay, ensemble size                        \\
\textbf{\texttt{dropout}}    & Dropout \citep{Gal2016Dropout}             &          $L_2$ decay, network, dropout rate                \\
\textbf{\texttt{bbb}}            & Bayes by Backprop  \citep{blundell2015weight}        &    prior mixture, network, early stopping                \\
\textbf{\texttt{hypermodel}}     & Hypermodel \citep{Dwaracherla2020Hypermodels} &                    $L_2$ decay, prior, bootstrap, index dimension \\
\textbf{\texttt{ensemble+}} & Ensemble + prior functions  \citep{osband2018rpf}  &      $L_2$ decay, ensemble size, prior scale, bootstrap                    \\
\textbf{\texttt{sgmcmc}}         & Stochastic Langevin  MCMC \citep{welling2011bayesian}  &               learning rate, prior, momentum           \\ \hline
\end{tabular}%
}
\end{center}
\vspace{-3mm}
\end{table}

\vspace{-1mm}
\section{Results}
\label{sec:results}
\vspace{-1mm}

We evaluate the benchmark agents of Section \ref{sec:agents} across the Neural Testbed.
We begin with an analysis of marginal predictions where, after agent tuning, all approaches are able to make reasonably good predictions.
However, when we examine \textit{joint} predictions we find that agent performance can vary drastically, even for well-tuned agents.
If an agent cannot output accurate joint predictions in the testbed, we should question if we expect that same agent to perform better other settings.
These results provide significant new insights to the the design of effective learning agents, and are a major contribution of this paper.

%%%%%%%%%%%%%%%%%%%%%%%%%%%%%%%%%%%%%%%%%%%%%%%%%%%%%%%%%%%%%%%%%%%%%%%%%%%%%% Marginals
\vspace{-1mm}
\subsection{Performance in marginal predictions}
\label{sec:results_marginal}
\vspace{-1mm}

% \1 This testbed is not entirely trivial, you need an MLP to learn to do well in it.
We begin our evaluation of benchmark approaches to Bayesian deep learning in marginal predictions ($\tau=1$).
One of the first questions one might consider is whether the generative model as outlined in Section~\ref{sec:generative} represents a meaningful challenge for deep learning systems.
Figure~\ref{fig:mlp-baseline} compares the performance of naive uniform class probabilities, logistic regression, and a tuned 2-layer MLP.
This simple comparison demonstrates that the Neural Testbed is not trivially solved by agents without deep learning architectures.

\begin{minipage}{.5\linewidth}
    % \vspace{-4mm}
    \begin{figure}[H]
    \vspace{-2mm}
    \centering
    \includegraphics[width=.8\linewidth]{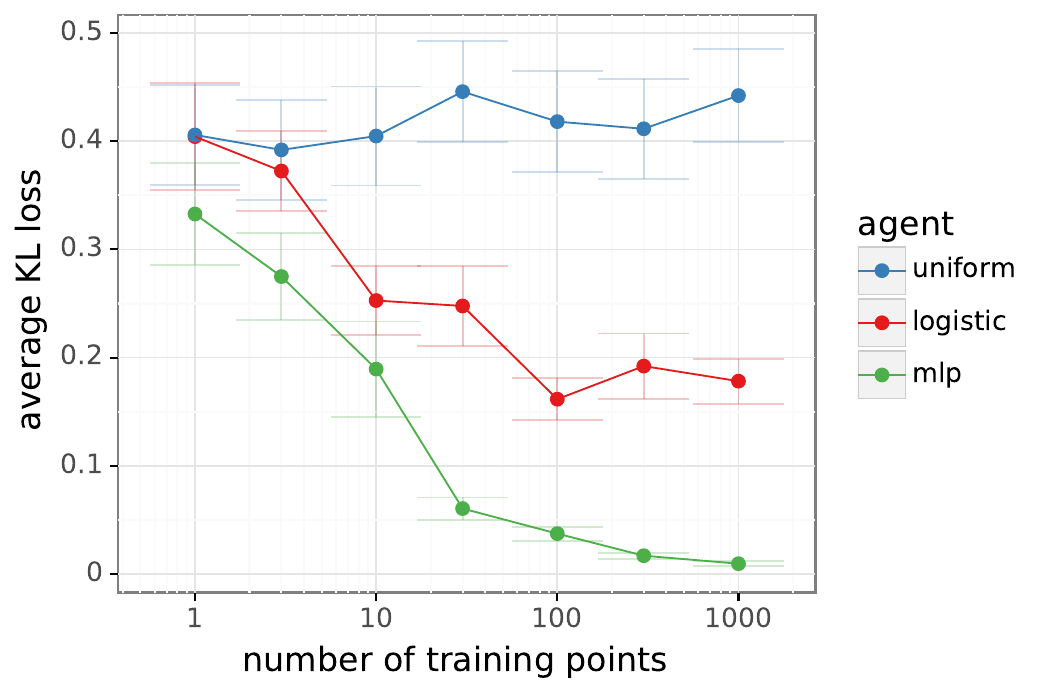}
    \vspace{-2mm}
    \caption{Performance with growing data.}
    \label{fig:mlp-baseline}
    \end{figure}
\end{minipage}
~
\begin{minipage}{0.45\linewidth}
    \begin{table}[H]
    % \vskip 0.15in
    \begin{center}
    \begin{small}
    \begin{sc}
    \resizebox{\columnwidth}{!}{%
    \begin{tabular}{lcccc}
    \toprule
          Agent &   Accuracy &    ECE   &   $\KL^1$ &  $\KL^{10}$  \\
    \midrule
            mlp &     0.793 &  0.078 &       0.129 &        1.367  \\
    % \midrule
       ensemble &     0.792 &  0.079 &       0.128 &        1.356  \\
        dropout &     0.793 &  0.080 &       0.128 &        1.347  \\
            bbb &     0.792 &  0.079 &       0.129 &        1.375  \\
     hypermodel &     0.793 &  0.081 &       0.130 &        \textbf{1.107}  \\
      ensemble+ &     0.790 &  0.085 &       0.129 &        \textbf{1.015}  \\
         sgmcmc &     0.796 &  0.082 &       0.122 &        \textbf{0.947}  \\
    \bottomrule
    \end{tabular}%
    }
    \end{sc}
    \end{small}
    \end{center}
    \caption{Agent performance, deviation from MLP greater than 2 stderr in bold.}
    \label{tab:ece}
    \vskip -0.02in
    \end{table}
\end{minipage}
% \vspace{-2mm}

% \1 Basically everything performs almost exactly the same... once you tune everything.
Marginal predictions have been the focus of the Bayesian deep learning literature.
Despite this focus, Figure~\ref{fig:overall_performance_both} shows that none of the benchmark methods significantly outperform a well-tuned MLP baseline in terms of $\KL^1$.
This observation is mirrored when we examine the average classification accuracy \textit{and} expected calibration error (ECE) across the testbed (Table~\ref{tab:ece}).
These results are different from the empirical observations in other challenge datasets, where much agent development has focused on improving ECE, and present an interesting new observation in the Bayesian deep learning literature \citep{nado2021uncertainty}.
We have two main hypothesis for this discrepancy.
First, our agents are tuned for $\KL^{\rm agg} = \KL^1 + \frac{1}{10}\KL^{10}$, not ECE (see Appendix \ref{app:agents}).
Second, the generative model of Section~\ref{sec:generative} matches the agent architecture, with inputs sampled i.i.d. $N(0,I)$.
Investigating the conditions in which these results hold more generally is an exciting area for future research.
\vspace{-1mm}
\subsection{Performance beyond marginals}
\label{sec:results_higher}
\vspace{-1mm}

One of the key contributions of this paper is to evaluate predictive distributions beyond marginals.
Figure~\ref{fig:overall_performance_both} shows that \texttt{sgmcmc} is the top-performing agent overall.
This should be reassuring to the Bayesian deep learning community and beyond.
In the limit of large compute this agent should recover the `gold-standard' of Bayesian inference, and it does indeed perform best \citep{welling2011bayesian}.
However, some of the most popular approaches in this field (\texttt{ensemble}, \texttt{dropout}) do not actually provide good approximations to the predictive distributions of order $\tau=10$.
In fact, we even see that \texttt{ensemble+} and \texttt{hypermodels} can provide much better approximations to the Bayesian posterior than `fully Bayesian' VI approaches like \texttt{bbb} \citep{wilson2020bayesian}.
We note too that while \texttt{sgmcmc} performs best, it also requires orders of magnitude more computation than competitive methods even in this toy setting (see Appendix \ref{app:testbed-speed}).
As we scale to more complex environments, it may therefore be worthwhile to consider alternative approaches.

\begin{figure}[!ht]
\centering
\hspace{-10mm}
    \begin{minipage}{0.65\linewidth}
    \centering
        \includegraphics[height=3.6cm]{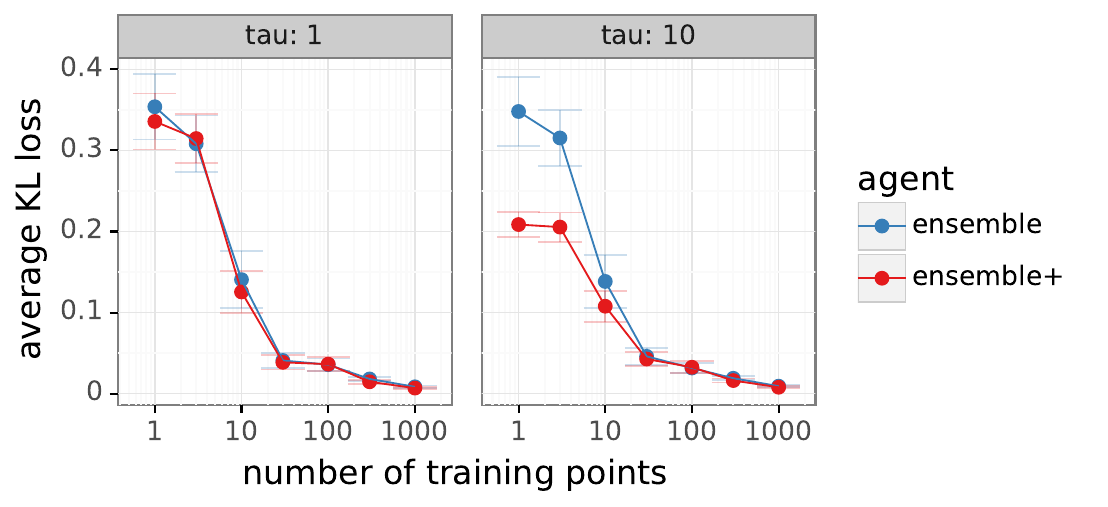}
        \vspace{-2mm}
        \caption{Prior functions help with joint predictions.}
        \label{fig:synthetic_ensemble_high_tau}
    \end{minipage}
    \hfill
    \begin{minipage}{0.37\linewidth}
    \centering
        \includegraphics[height=3.6cm]{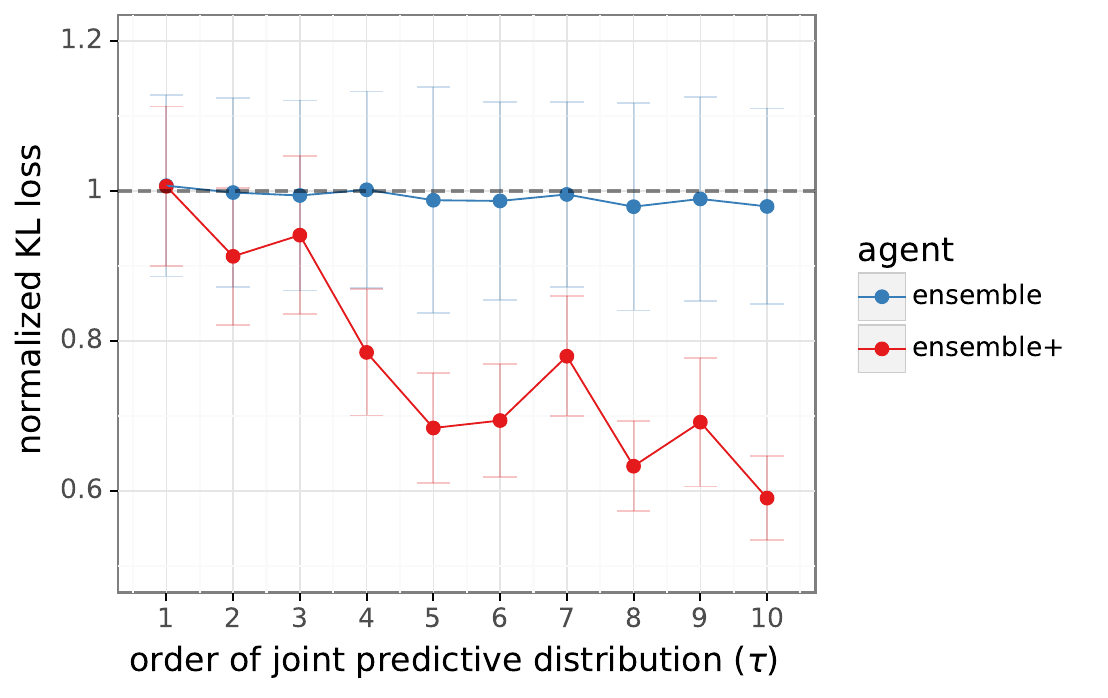}
        \vspace{-5mm}
        \caption{Benefit grows with $\tau$.}
        \label{fig:tau_robustness}
    \end{minipage}
\end{figure}

To see where some agents are able to outperform, we compare \texttt{ensemble} and \texttt{ensemble+} under the medium SNR regime.
These agents are identical, except for the addition of a randomized prior function \citep{osband2018rpf}.
Figure~\ref{fig:synthetic_ensemble_high_tau} shows that, although these methods perform similarly in the quality of their marginal predictions ($\tau=1$), the addition of a prior function greatly improves the quality of joint predictive distributions ($\tau=10$) in the low data regime.
Note that, since the testbed considers 2D inputs, $100$ training points may already be considered as in the high data regime.
% As expected, the effects of a prior function diminish as the agent gathers more data from the environment, so that the methods perform similarly in high data regimes.
% These results are quite intuitive: the effects of a prior function diminish as the agent gathers more data from the environment, so that the methods perform similarly in high data regimes.
Figure~\ref{fig:tau_robustness} provides some insight for how this benefit scales with the order $\tau$ of the predictive distribution.
We can see a clear trend that as $\tau$ increases so does the separation between agents \texttt{ensemble} and \texttt{ensemble+}.
For more intuition on \textit{how} prior functions are able to drive this benefit, see Appendix~\ref{app:testbed-visualize}.

% \begin{figure}[!ht]
%     \centering
%     \begin{subfigure}[b]{0.65\textwidth}
%         \centering
%         \includegraphics[width=.95\linewidth]{figures/synthetic_ensemble_high_tau.pdf}
%         \caption{Additive prior functions help with joint predictions.}
%         \label{fig:synthetic_ensemble_high_tau}
%     \end{subfigure}%
%     ~ 
%     \begin{subfigure}[b]{0.35\textwidth}
%         \centering
%         \includegraphics[width=.99\linewidth]{figures/tau_sensitivity_2.pdf}
%         \vspace{-3mm}
%         \caption{Prior helps more as $\tau$ grows.}
%         \label{fig:tau_robustness}
%     \end{subfigure}
%     \caption{Caption place holder}
%     \label{fig:tau}
% \end{figure}

\vspace{-1mm}

\section{Sequential decisions}
\label{sec:sequential}
\vspace{-1mm}

% \begin{outline}
% \1 This is a section to describe the sequential decision problems we derive from the testbed.
%     \2 Simple bandit problem made from the testbed.
%     \2 The agents that do well on it are the ones that have good high tau predictions.
% \end{outline}

% Section \ref{sec:results} outlines agent performance on the Neural Testbed.
% We see that several state-of-the-art agents that perform similarly in marginal prediction have markedly different quality in terms of joint distributions.
In this section, we will form a sequential decision problem based on the Neural Testbed, and show that it is the quality in \textit{joint} predictions that is essential to driving good performance in sequential decision making.
Further, we show that the insights gained from the simple 2D Neural Testbed can extend to high-dimensional decision problems.

%%%%%%%%%%%%%%%%%%%%%%%%%%%%%%%%%%%%%%%%%%%%%%%%%%%%%%%%%%%%%%%%%%%%%%%%%%%%%% Bandit problem
\vspace{-1mm}
\subsection{Neural bandit}
\label{sec:bandit_problem}
\vspace{-1mm}

% One of the main motivations for the Neural Testbed is to design agents suitable for sequential decision problems.
% In this section, we outline an empirical evaluation of predictive distributions for sequential decisions through a \textit{Thompson sampling} (TS) agent \citep{Thompson1933}.

We use the generative model of the Neural Testbed to define a class of bandit problems \citep{gittins1979bandit}.
First, we sample a set of $N$ actions $\mathcal{X} = \{x_1, \dots, x_N\}$ i.i.d. from a $d$-dimensional standard normal distribution. We then sample an environment $\environment$, which specifies the conditional probability $\environment(Y_{t+1} \in \cdot | X_t)$, according to the class of generative models described in Section~\ref{sec:generative}.
We pick the temperature, which controls the SNR, to be $0.1$.
At each timestep $t$, an agent selects an action $X_t \in \mathcal{X}$ and receives a reward $R_{t+1} = Y_{t+1}$. Let $\overline{R}_{x} = \Exp\left[R_{t+1} | \environment, X_t=x \right]$ denote the expected reward of action $x$ conditioned on the environment, and let $X_* = \arg\max_{x \in \mathcal{X}} \overline{R}_{x}$ denote the optimal action. We assess agent performance through ${\tt regret}(T) := \sum_{t=0}^{T-1} \Exp \left[ \overline{R}_{X_*}- \overline{R}_{X_t} \right]$, which measures the shortfall in expected cumulative rewards relative to an optimal decision maker.

% First, we sample actions $\Ac = \{a^1, .., a^A\}$ i.i.d. from $N(0, I)$ and a probability function $f^*$ determined by a random 2-layer ReLU MLP according to the genrative model in Section~\ref{sec:generative}.
% At each timestep $t$ an agent selects action $A_t \in \Ac$, and receives a stochastic reward $R_{t+1}$ with $\Prob (R_{t+1}=1) = f^*(A_t)$. We assess agent performance through the ${\tt regret}(T) := \sum_{t=0}^{T-1} \Exp \left[\max_a f^*(a) - f^*(A_t) \right]$, which measures the shortfall in expected cumulative rewards relative to an optimal decision maker.

We evaluate the testbed agents on these bandit problems through actions selected by Thompson sampling, varying only the posterior predictive distributions that TS samples from.
A TS agent requires an approximate posterior distribution over the environment, which is supplied by the testbed agents.
At each timestep, TS samples an environment from the approximate posterior and selects an action that optimizes for the sampled environment \citep{Thompson1933, russo2017tutorial}.
% This simple heuristic can effectively balance the needs of exploration and exploitation, and allows us to evaluate the ability of an approximate posterior to drive decision making.
A complete algorithm is presented in Appendix \ref{app:sequential}.

%%%%%%%%%%%%%%%%%%%%%%%%%%%%%%%%%%%%%%%%%%%%%%%%%%%%%%%%%%%%%%%%%%%%%%%%%%%%%% Bandit results
\vspace{-1mm}
\subsection{Agent performance}
\label{sec:bandit_performance}
\vspace{-1mm}

% \begin{figure*}
% \centering
% \begin{minipage}[b]{.3\textwidth}
% \includegraphics[width=.99\linewidth]{figures/bandit_regret.pdf}
% \caption{Caption}\label{label-a}
% \end{minipage}\qquad
% \begin{minipage}[b]{.3\textwidth}
% \includegraphics[width=.99\linewidth]{figures/bandit_corr_tau_1.pdf}
% \caption{Caption}\label{label-b}
% \end{minipage}\qquad
% \begin{minipage}[b]{.3\textwidth}
% \includegraphics[width=.99\linewidth]{figures/bandit_corr_tau_10.pdf}
% \caption{Caption}\label{label-b}
% \end{minipage}
% \end{figure*}

\begin{figure}[!ht]
  \centering
  \includegraphics[width=.6\linewidth]{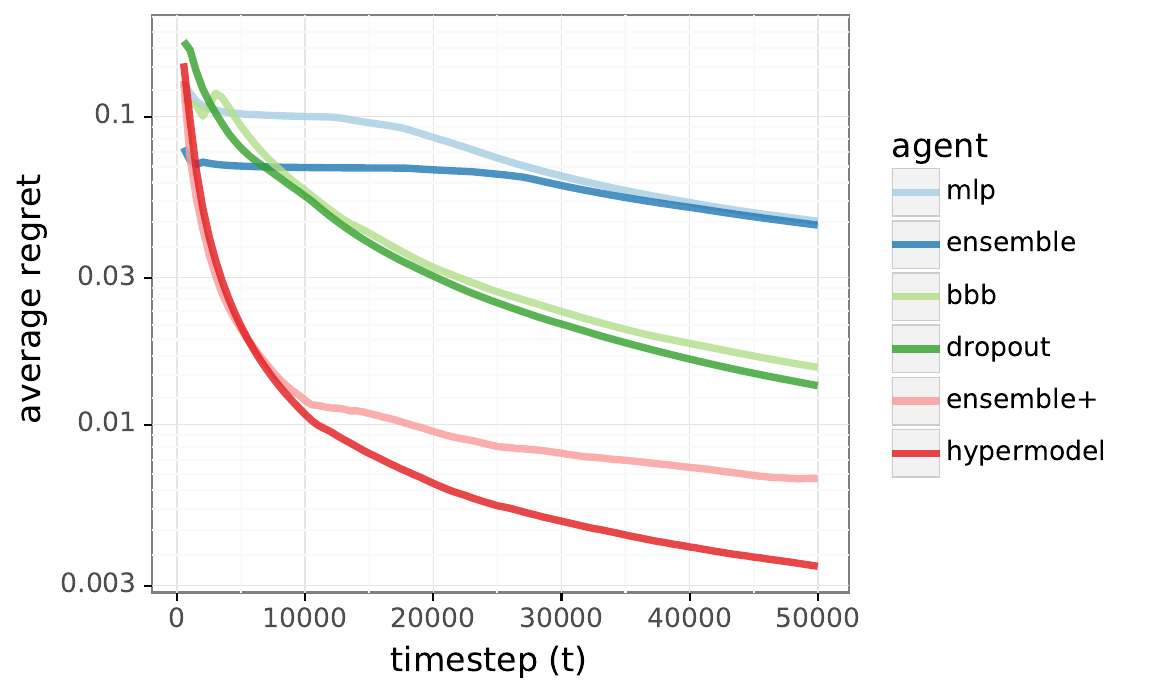}
  \vspace{-3mm}
  \caption{Learning agent impacts TS regret in neural bandits.}
\label{fig:bandit_regret_curve}
\vspace{-2mm}
\end{figure}

% We run a bunch of agents and all of them can learn in this problem.
We present empirical results of testbed agents on these random bandit problems with $N = 1000$ actions drawn from a $d = 50$ dimensional space.
Figure~\ref{fig:bandit_regret_curve} shows the average regret through time for each of the agents as selected by the Neural Testbed, averaged over 20 random seeds.\footnote{We omit \texttt{sgmcmc} as the computational demands are several orders of magnitude too large to consider in online learning.}
We can see that for each learning agent, the quality of decisions improves through time.
However, the quality of decisions is greatly affected by the choice of agent.
% As expected, the agents that perform best in sequential decision problems seem to correlate with those that make accurate \textit{joint} predictions on the testbed \citep{lu2021BeyondMarginal}.

\begin{figure}[!ht]
    \begin{minipage}{0.48\linewidth}
    \centering
    \includegraphics[width=.99\linewidth]{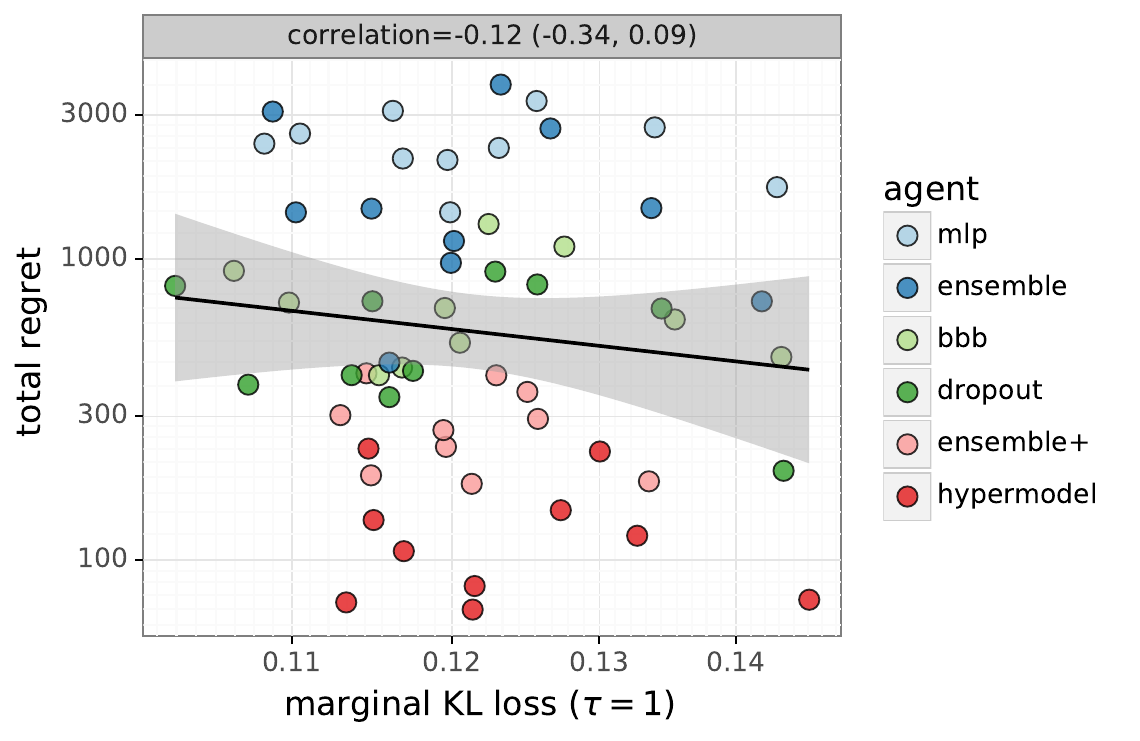}
    \vspace{-3mm}
    \caption{Testbed marginal performance is not significantly correlated with regret.}
    \vspace{-1mm}
    \label{fig:bandit_tau_1}
    \end{minipage}
    ~
    \begin{minipage}{0.48\linewidth}
    \centering
    \includegraphics[width=.99\linewidth]{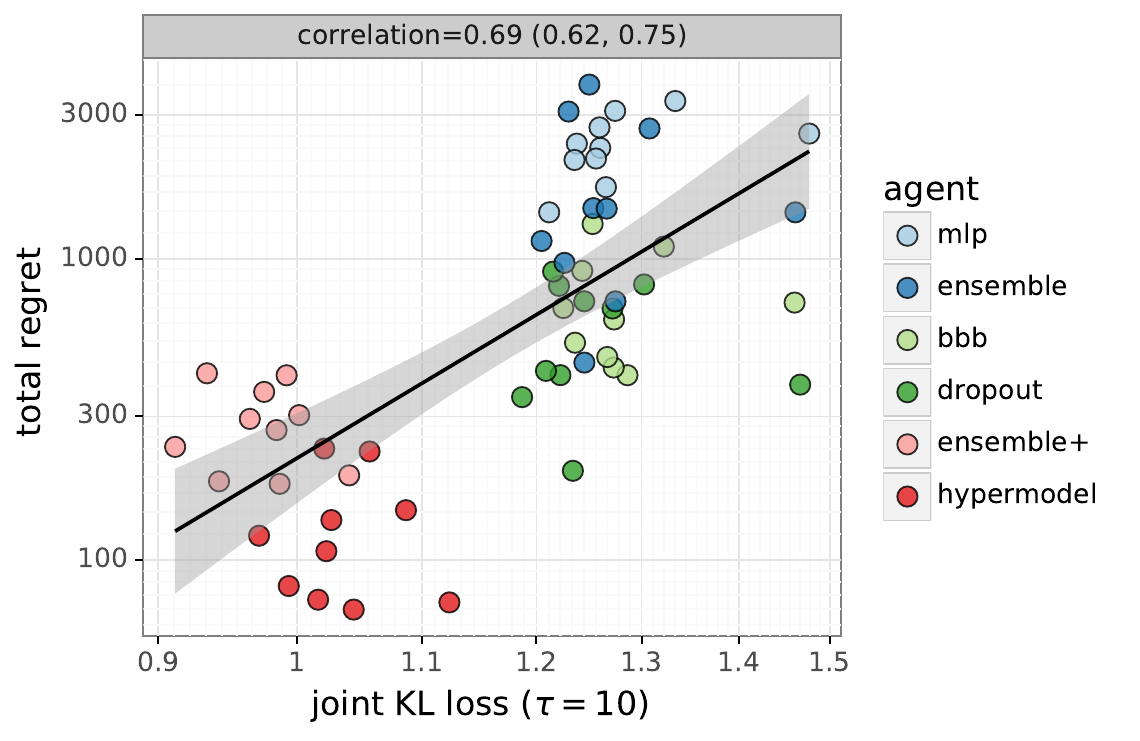}
    \vspace{-3mm}
    \caption{Testbed joint performance is highly correlated with regret.}
    \vspace{-1mm}
    \label{fig:bandit_tau_10}
    \end{minipage}
\end{figure}

% Generally, KL with tau=10 is a much better predictor of an agent's performance on sequential decision problems than KL with tau=1.
To investigate the relationship between predictions and decisions we repeat the experiment of Figure~\ref{fig:bandit_regret_curve} with 10 independent random initializations over all the testbed and bandit problems.
We then empirically investigate the correlation between $\KL^\tau$ and total regret at $T=50,000$ for both $\tau=1$ and $\tau=10$.
We use bootstrap sampling to estimate confidence intervals on the correlation coefficient on a logarithmic scale at the 5th and 95th percentiles.
Figures~\ref{fig:bandit_tau_1} and \ref{fig:bandit_tau_10} support our claim that performance in $\KL^{10}$ is highly correlated with performance in sequential decision problems, whereas correlation to marginals is not significant.
We would not expect a perfect correlation as the particular TS action selection strategy may introduce confounding factors, together with natural variability in seeds.

% These results are significant in that they scale to high input dimension and robust to input_dim.
% These results are significant in two ways.
% First, they provide empirical evidence that practical deep learning approaches separated by the quality of their \textit{joint} predictions, but not their marginals, can lead to differing performance in downstream tasks.
% Second, we show that our simple 2D testbed can provide insights that scale to much higher dimension problems.
% In Appendix~\ref{app:sequential_results} we show that these results are largely insensitive to input dimension.
% As such, we believe that sanity-checking approaches to probabilistic inference on the Neural Testbed offers a valuable complement to existing high-dimensional challenges in Bayesian deep learning.

% \begin{figure}[!ht]
%   \centering
%   \includegraphics[width=.9\linewidth]{figures/bandit_corr_tau_10.pdf}
%   \vspace{-3mm}
% \caption{Testbed joint performance $\tau=10$ is highly correlated with sequential decision performance.}
% \vspace{-1mm}
% \label{fig:bandit_tau_10}
% \end{figure}

%%%%%%%%%%%%%%%%%%%%%%%%%%%%%%%%%%%%%%%%%%%%%%%%%%%%%%%%%%%%%%%%%%%%%%%%%%%%%%%%%%%%%%%%%%% ROBUSTNESS
%%%%%%%%%%%%%%%%%%%%%%%%%%%%%%%%%%%%%%%%%%%%%%%%%%%%%%%%%%%%%%%%%%%%%%%%%%%%%%%%%%%%%%%%%%%
\section{Robustness of generative model}
\label{sec:robustness}

% We're going to have a bunch of experiments varying the kernel.
% Explain what the experiments are going to be and what they are going to show.
% \fillpara
The experiments of Sections~\ref{sec:results} and \ref{sec:sequential} are all performed with the generative model as described in Section~\ref{sec:generative}.
One natural concern is that these results might be sensitive to this choice of model, and so be less transferable to general deep learning research.
In this section we repeat these analyses under different generative models.
We find that the quality of \textit{joint} predictions and bandit performance is extremely robust across choice of generative models.

\begin{figure}[!th]
  \centering
  \vspace{-2mm}
  \includegraphics[width=0.95\linewidth]{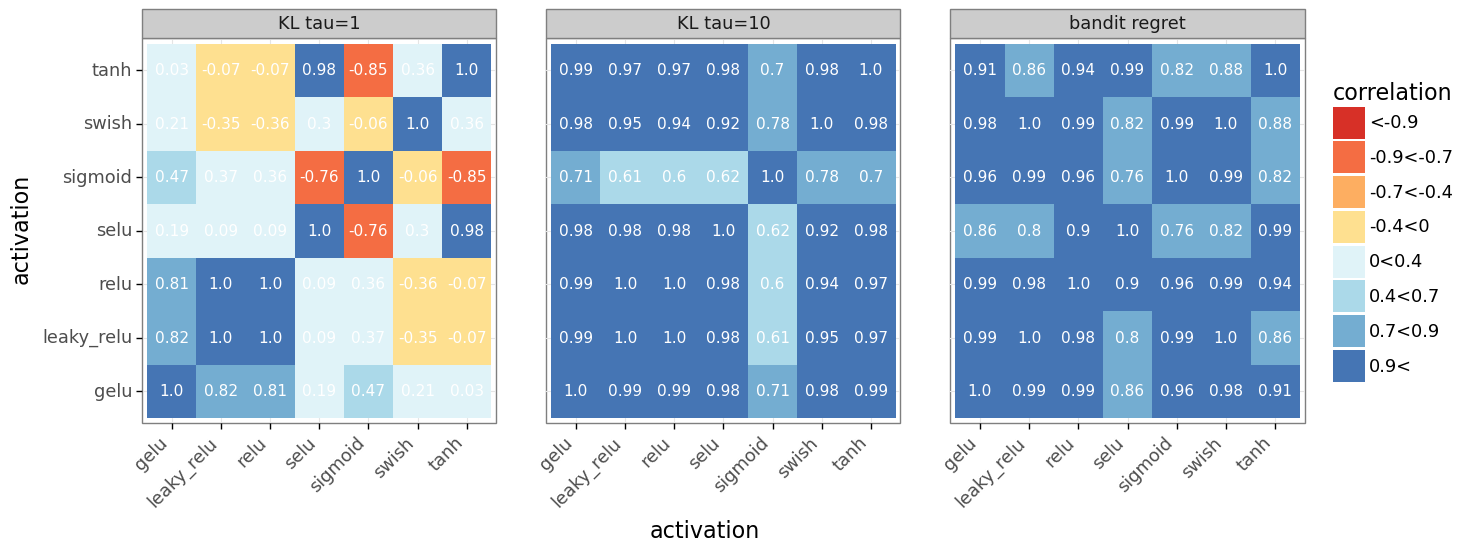}
  \vspace{-2mm}
  \caption{Correlation of agent performance across different activation functions.}
\label{fig:robustness_activation}
\end{figure}

\begin{figure}[!th]
  \centering
  \vspace{-2mm}
  \includegraphics[width=0.95\linewidth]{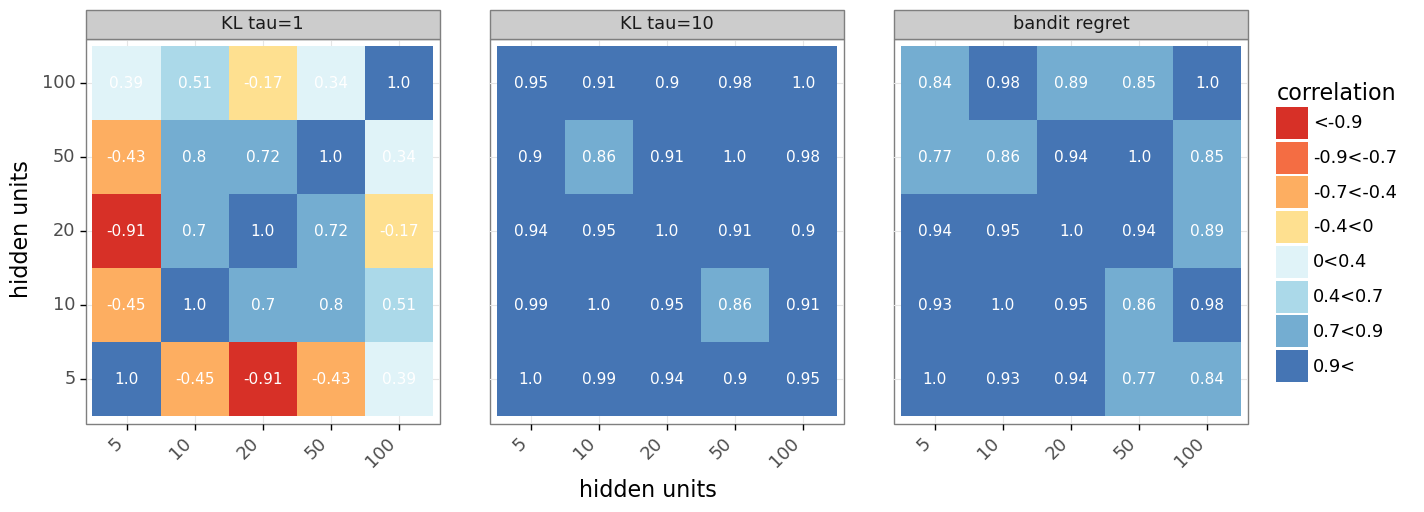}
  \vspace{-2mm}
  \caption{Correlation of agent performance across different hidden units.}
\label{fig:robustness_hidden}
\end{figure}

% Explain the methodology
For these experiments we take the tuned agents of Section~\ref{sec:results} and then evaluate these agents under different generative models.
Whereas these agent hyperparameters were tuned for the 2-layer ReLU MLP with 50-50 hidden units, we will also these agents over alternative environments varying:
\begin{itemize}[noitemsep, nolistsep]
    \item \textbf{activation}=[tanh, swish, sigmoid, selu, relu, leaky relu, gelu] (Figure~\ref{fig:robustness_activation}).
    \item \textbf{hidden units}=[5, 10, 20, 50, 100] (Figure~\ref{fig:robustness_hidden}).
\end{itemize}
Evaluation for each of these environments $\Ec_i$ proceeds as before: the agent is trained on data generated by $\Ec_i$ and then evaluated on the quality of predictions on testing data from $\Ec_i$.
If the qualitative results under different environments are similar, then we know that our results are somewhat robust to the exact generative model we choose.
% In each case the agents still get to train on the observed data, but the generative model no longer matches what they were tuned on.

Figure~\ref{fig:robustness_activation} and \ref{fig:robustness_hidden} examine the empirical correlation coefficient between the vector of agent evaluations, under the metrics $\KL^1, \KL^{10}$ and bandit regret.
We see that, the marginal evaluations are highly correlated for `similar' generative models (e.g. ReLU and leaky ReLU) but can even be anti-correlated when the models stray too far.
However, the correlations are very high across a wide range of generative models when we look at either the quality of joint predictions or the regret in the bandit problems.
These results help to build confidence in the key observations we make in this paper.
Notably, they suggest that the separation of agents in terms of performance on joint prediction (Figure~\ref{fig:overall_performance_both}) is not too sensitive to the choice of generative model, and so may hold some wider insight relevant to the community.
Follow up work has confirmed that these results are also highly correlated with performance on benchmark datasets \citep{osband2022evaluating}.

%%%%%%%%%%%%%%%%%%%%%%%%%%%%%%%%%%%%%%%%%%%%%%%%%%%%%%%%%%%%%%%%%%%%%%%%%%%%%%%%%%%%%%%%%%% CONCLUSION
%%%%%%%%%%%%%%%%%%%%%%%%%%%%%%%%%%%%%%%%%%%%%%%%%%%%%%%%%%%%%%%%%%%%%%%%%%%%%%%%%%%%%%%%%%%
\section{Conclusion}
\label{sec:conclusion}

% Introduce the Neural Testbed and opensource it.
% This paper introduces the Neural Testbed, and the associated opensource release at \github.
The Neural Testbed investigates the quality of predictive uncertainty in joint predictions, as well as marginals.
With this simple and clear 2D challenge we aim to build understanding that can inform the field's wider efforts in deep learning.
We have shown that results on the testbed can offer new insights to agent development.
Further, we establish that the insights gained in the testbed can scale up to complex and high-dimensional decision problems.

% This is just the beginning of work, not the end.
Beyond the results in this paper, we believe this work can provide a base for future research:
% \vspace{-1mm}
\begin{itemize}[noitemsep, nolistsep]
    \item Can we design better learning algorithms for joint predictions, as well as marginals?
    % \item Can this joint perspective improve agent robustness?
    % \item Can good joint predictions facilitate streaming agents?
    \item Are there analogous results to Figure~\ref{fig:overall_performance_both} on large-scale challenge datasets?
    \item How can effective joint predictions drive better decisions?
\end{itemize}
% \vspace{-1mm}
We believe that studying these simple testbed problems can help foster interplay between theory and practice, improve accessibility in the field, and complement existing research.
We hope that this will accelerate the growth of \textit{understanding} in the field and, ultimately, drive forward the design of better learning agents.
{
\small
\bibliographystyle{apalike}
\bibliography{references}
}

\appendix
\newpage

%%%%%%%%%%%%%%%%%%%%%%%%%%%%%%%%%%%%%%%%%%%%%%%%%%%%%%%%%%%%%%%%%%%%%%%%%%%%%%%% CODE
%%%%%%%%%%%%%%%%%%%%%%%%%%%%%%%%%%%%%%%%%%%%%%%%%%%%%%%%%%%%%%%%%%%%%%%%%%%%%%%%
\section{Open source code}
\label{app:code}

This section is meant to give an overview of our opensource code.
Together with our paper submission we include a link to anonymous github repository.
{
\begin{itemize}[leftmargin=*]
    \item \mbox{\textbf{\texttt{neural\_testbed}}: \githubtestbed}
\end{itemize}
}
Together with this git repo, we include a `tutorial colab' -- a Jupyter notebooks that can be run in the browser without requiring any local installation at \texttt{neural\_testbed/tutorial.ipynb}.
Our library is written in Python, and relies heavily on JAX for scientific computing \citep{jax2018github}.
We view this open-source effort as a major contribution of our paper.

%%%%%%%%%%%%%%%%%%%%%%%%%%%%%%%%%%%%%%%%%%%%%%%%%%%%%%%%%%%%%%%%% PSEUDOCODE
%%%%%%%%%%%%%%%%%%%%%%%%%%%%%%%%%%%%%%%%%%%%%%%%%%%%%%%%%%%%%%%%%
%%%%%%%%%%%%%%%%%%%%%%%%%%%%%%%%%%%%%%%%%%%%%%%%%%%%%%%%%%%%%%%%%%%%%%%%%%%%%%%% PSEUDOCODE
%%%%%%%%%%%%%%%%%%%%%%%%%%%%%%%%%%%%%%%%%%%%%%%%%%%%%%%%%%%%%%%%%%%%%%%%%%%%%%%%
\section{Testbed Pseudocode}
\label{app:testbed_pseudo_code}

We present the testbed pseudocode in this section. Specifically, Algorithm~\ref{alg:neural-testbed} is the pseudocode for our neural testbed, and Algorithm~\ref{alg:mc-estimation} is an approach to estimate the likelihood of a test data $\tau$-sample conditioned on an agent's belief, based on the standard Monte-Carlo estimation.
%
% and Algorithm~\ref{alg:rp-estimation} are two different approaches to estimate the likelihood of a test data $\tau$-sample conditioned on an agent's belief. Algorithm~\ref{alg:mc-estimation} is based on the standard Monte-Carlo estimation, while Algorithm~\ref{alg:rp-estimation} adopts a random partitioning approach. 
% As a rule of thumb, we recommend the users to use Algorithm~\ref{alg:mc-estimation} for $\tau < 10$ and Algorithm~\ref{alg:rp-estimation}  with a feature dimension between $5$ and $10$ for $\tau \geq 10$.
The presented testbed pseudocode works for any prior $\Prob(\environment \in \cdot)$ over the environment and any input distribution $P_X$, including the ones described in Section \ref{sec:generative}.
We also release full code and implementations in Appendix~\ref{app:code}.

% We present the testbed pseudocode in this section. Specifically, Algorithm~\ref{alg:neural-testbed} is the pseudocode for our neural testbed, and Algorithm~\ref{alg:mc-estimation} and Algorithm~\ref{alg:rp-estimation} are two different approaches to estimate the likelihood of a test data $\tau$-sample conditioned on an agent's belief. Algorithm~\ref{alg:mc-estimation} is based on the standard Monte-Carlo estimation, while Algorithm~\ref{alg:rp-estimation} adopts a random partitioning approach. 
% % As a rule of thumb, we recommend the users to use Algorithm~\ref{alg:mc-estimation} for $\tau < 10$ and Algorithm~\ref{alg:rp-estimation}  with a feature dimension between $5$ and $10$ for $\tau \geq 10$.
% The presented testbed pseudocode works for any prior $\Prob(\environment \in \cdot)$ over the environment and any input distribution $P_X$, including the ones described in Section \ref{sec:generative}.
% We also release full code and implementations at \github.
% we also discuss some core technical issues in the neural testbed design. Specifically,
% % Appendix~\ref{app:likelihood_estimation} discusses how to estimate the likelihood of an agent's belief distribution;
% Appendix~\ref{app:evaluation_real_data} discusses how to extend the testbed to agent evaluation on real data and Appendix~\ref{app:experiment_parameters} explains our choices of experiment parameters.

\begin{algorithm}
\caption{Neural Testbed}
\label{alg:neural-testbed}
% \setstretch{1.1}
\begin{algorithmic}
\REQUIRE the testbed requires the following inputs
\begin{enumerate}
\item prior distribution over the environment $\Prob(\environment \in \cdot)$, input distribution $P_X$
\item agent $f_\theta$
\item number of training data $T$, test distribution order $\tau$
\item number of sampled problems $J$, number of test data samples $N$
\item parameters for agent likelihood estimation, as is specified in Algorithm~\ref{alg:mc-estimation} 
% and \ref{alg:rp-estimation}
\end{enumerate}
\vspace{0.3em}
\FOR{$j=1, 2, \ldots, J$}
\vspace{0.3em}
\STATE \textbf{Step 1: sample environment and training data}
    \STATE \hspace{1em} 1. sample environment $\environment \sim \Prob(\environment \in \cdot)$
    \STATE \hspace{1em} 2. sample $T$ inputs $X_0, X_1, \ldots, X_{T-1}$ i.i.d. from $P_X$ \STATE \hspace{1em} 3. sample the training labels
    $Y_1, \ldots, Y_T$ conditionally i.i.d. as
    \[
    Y_{t+1} \sim \Prob \left(Y \in \cdot \middle | \environment, X = X_t \right) \quad \forall t=0, 1, \ldots, T-1
    \]
    \STATE \hspace{1em} 4. choose the training dataset as
    $
    \data_T = \left \{ \left(X_t, Y_{t+1} \right), \, t=0, \ldots, T-1 \right \}$
\vspace{0.3em}
\STATE \textbf{Step 2: train agent}
    \STATE \hspace{1em} train agent $f_{\theta_T}$ based on training dataset $\data_T$
\vspace{0.3em}
\STATE \textbf{Step 3: compute likelihoods}
    \STATE \hspace{1em} \textbf{for} $n=1,2,\ldots, N$ \textbf{do}
        \STATE \hspace{2em} 1. sample $X^{(n)}_{T}, \ldots, X^{(n)}_{T+\tau-1}$ i.i.d. from $P_X$
        \STATE \hspace{2em} 2. generate $Y^{(n)}_{T+1}, \ldots, Y^{(n)}_{T+\tau}$ conditionally independently as
    \[
    Y_{t+1}^{(n)} \sim \Prob \left(Y \in \cdot \middle | \environment, X = X_t^{(n)} \right) \quad \forall t=T, T+1, \ldots, T+\tau-1
    \]
    % the testing dataset is $\tilde{\data}^{(n)}_\tau = \left \{
    % \left ( X_t^{(n)}, Y_{t+1}^{(n)} \right), t=T, \ldots, T+\tau-1
    % \right \}$
    \STATE \hspace{2em} 3. compute the likelihood under the environment $\environment$ as
    \begin{align*}
        \textstyle p_{j,n} = \Prob \left(
        Y_{T+1:T+\tau}^{(n)}\middle | \environment, 
         X_{T:T+\tau-1}^{(n)}
        \right) = \prod_{t=T}^{T+\tau-1} 
        \Pr \left(Y_{t+1}^{(n)} \middle | \environment, X_{t}^{(n)}
        \right)
    \end{align*}
    \STATE \hspace{2em} 4. estimate the likelihood conditioned on the agent's belief 
    \[
    \hat{p}_{j,n} \approx \Prob \left(\hat{Y}_{T+1:T+\tau} = Y_{T+1:T+\tau}^{(n)} \middle | \theta_T,   X_{T:T+\tau-1}^{(n)},  Y_{T+1:T+\tau}^{(n)}  \right) ,\] 
    \STATE \hspace{2em} based on Algorithm~\ref{alg:mc-estimation} 
    % or \ref{alg:rp-estimation} 
    with test data $\tau$-sample 
    $\left(X_{T:T+\tau-1}^{(n)}, Y_{T+1:T+\tau}^{(n)} \right)$. 
    \vspace{0.3em}
\ENDFOR \\
\RETURN $\frac{1}{JN} \sum_{j=1}^J \sum_{n=1}^N \log \left( p_{j,n}/\hat{p}_{j,n} \right)$
\end{algorithmic}
% \setstretch{1}
\end{algorithm}

\begin{algorithm}
\caption{Monte Carlo Estimation of Likelihood of Agent's Belief}
\label{alg:mc-estimation}
\begin{algorithmic}
\REQUIRE the Monte-Carlo estimation requires the following inputs
\begin{enumerate}
    \item trained agent $f_{\theta_T}$ and number of Monte Carlo samples $M$
    \item test data $\tau$-sample $\left(X_{T:T+\tau-1}, Y_{T+1:T+\tau} \right)$
\end{enumerate}
\STATE \textbf{Step 1:} sample $M$ models $\hat{\environment}_1, \ldots, \hat{\environment}_M$ conditionally i.i.d. from $\Prob \left( \hat{\environment} \in \cdot \middle | \theta_T \right)$
\STATE \textbf{Step 2:} estimate $\hat{p}$ as
\[
\hat{p} = \frac{1}{M} \sum_{m=1}^M
\Prob \left(\hat{Y}_{T+1:T+\tau} = Y_{T+1:T+\tau} \middle | \hat{\environment}_m,   X_{T:T+\tau-1},  Y_{T+1:T+\tau}  \right)
\]
\RETURN $\hat{p}$
\end{algorithmic}
\end{algorithm}

In addition to presenting the testbed pseudocode, we also explain our choices of experiment parameters in Appendix~\ref{app:agents}.
To apply Algorithm \ref{alg:neural-testbed}, we need to specify an input distribution $P_X$ and a prior distribution on the environment $\Prob(\environment \in \cdot)$. Recall from Section \ref{sec:generative} that we consider binary classification problems with input dimension $2$. We choose $P_X = N(0, I)$, and we consider three environment priors distinguished by a temperature parameter that controls the signal-to-noise ratio (SNR) regime. We sweep over temperatures in $\{0.01, 0.1, 0.5\}$. The prior distribution $\Prob(\environment \in \cdot)$ is induced by a distribution over MLPs with 2 hidden layers and ReLU activation. The MLP is distributed according to standard Xavier initialization, except that biases in the first layer are drawn from $N(0, \frac{1}{2})$. The MLP outputs two units, which are divided by the temperature parameter and passed through the softmax function to produce class probabilities.
The implementation of this generative model is in our open source code under the path \url{/generative/factories.py}.

We now describe the other parameters we use in the Testbed. In Algorithm \ref{alg:neural-testbed}, we pick the order of predictive distributions $\tau \in \{1, 10\}$, training dataset size $T \in \{1, 3, 10, 30, 100, 300, 1000\}$, number of sampled problems $J = 10$, and number of testing data $\tau$-samples $N = 1000$. To apply Algorithm~\ref{alg:mc-estimation}, we sample 
$M = 1000$ models from the agent.

%%%%%%%%%%%%%%%%%%%%%%%%%%%%%%%%%%%%%%%%%%%%%%%%%%%%%%%%%%%%%%%%% AGENTS
%%%%%%%%%%%%%%%%%%%%%%%%%%%%%%%%%%%%%%%%%%%%%%%%%%%%%%%%%%%%%%%%%
\section{Agents}
\label{app:agents}

% In this section, we describe the implementation of the agents we discuss in Section \ref{sec:agents} namely: mlp, ensemble, dropout, Bayes by backprop, stochastic Langevin MCMC, ensemble+ and hypermodel.

In this section, we describe the benchmark agents in Section~\ref{sec:agents} and the choice of various hyperparameters used in the implementation of these agents.
The list of agents include MLP, ensemble, dropout, Bayes by backprop, stochastic Langevin MCMC, ensemble+ and hypermodel.
We will also include other agents such as KNN, random forest, and deep kernel, but the performance of these agents was worse than the other benchmark agents, so we chose not to include them in the comparison in Section~\ref{sec:results}.
In each case, we attempt to match the ``canonical'' implementation. The complete implementation of these agents including the hyperparameter sweeps used for the Testbed are available in Appendix~\ref{app:code}.
We make use of the Epistemic Neural Networks notation from  \citep{osband2021epistemic} in our code. 
We set the default hyperparameters of each agent to be the ones that minimize the aggregated KL score $\KL^{\rm agg} = \KL^1 + \frac{1}{10}\KL^{10}$.

%%%%%%%%%%%%%%%%%%%%%%%%%%%%%%%%%%%%%%%%%%%%%%%%%%%%%%%%%%%%%%% MLP
\subsection{MLP}
The \texttt{mlp} agent learns a 2-layer MLP with 50 hidden units in each layer by minimizing the cross-entropy loss with $L_2$ weight regularization. The $L_2$ weight decay scale is chosen either to be $\lambda\frac{1}{T}$ or $\lambda \frac{d \sqrt{\beta}}{T}$, where $d$ is the input dimension, $\beta$ is the temperature of the generative process and $T$ is the size of the training dataset. We sweep over $\lambda \in \{10^{-4}, 10^{-3}, 10^{-2}, 10^{-1}, 1, 10, 100\}$. We implement the MLP agent as a special case of a deep ensemble (\ref{sec:agents-vanila_ensemble}). The implementation and hyperparameter sweeps for the \texttt{mlp} agent can be found in our open source code, as a special case of the \texttt{ensemble} agent, under the path \url{/agents/factories/ensemble.py}. 
%\zheng{$|\data_T|=T$, this might simplify the notation. Also, maybe we should change all ``L2" to $L_2$.}

%%%%%%%%%%%%%%%%%%%%%%%%%%%%%%%%%%%%%%%%%%%%%%%%%%%%%%%%%%%%%%%%% Vanilla ensemble
\subsection{Ensemble}
\label{sec:agents-vanila_ensemble}
We implement the basic ``deep ensembles'' approach for posterior approximation \citep{lakshminarayanan2017simple}. The \texttt{ensemble} agent learns an ensemble of MLPs by minimizing the cross-entropy loss with $L_2$ weight regularization. The only difference between the ensemble members is their independently initialized network weights. We chose the $L_2$ weight scale to be either  $\lambda\frac{1}{M T}$ or $\lambda \frac{d \sqrt{\beta}}{M T}$, where $M$ is the ensemble size, $d$ is the input dimension, $\beta$ is the temperature of the generative process, and $T$ is the size of the training dataset. We sweep over ensemble size $M \in \{1, 3, 10, 30, 100\}$ and $\lambda \in \{10^{-4}, 10^{-3}, 10^{-2}, 10^{-1}, 1, 10, 100\}$.
We find that larger ensembles work better, but this effect is within margin of error after 10 elements.
%\morteza{Maybe add the plots showing this result.}
The implementation and hyperparameter sweeps for the \texttt{ensemble} agent can be found in our open source code under the path \url{/agents/factories/ensemble.py}. 

% \zheng{optional: in Appendix B, we use $M$ to denote something different. would be better to use a different notation.}
 
% We use the same loss function and sampled minibatch for all particles  of the ensemble. The only difference between the particles is their random parameter initialization.
% We use the regularized negative log likelihood as the loss
% \[
% \Lc (\theta, \tilde{\data}) = - \frac{1}{ |\tilde{\data}|} \sum_{(x, y) \in \tilde{\data}} \frac{1}{|Z|}\sum_{z \in Z} \mathrm{XEntropy} \left( y, f_\theta(x, z) \right) +
% \frac{\tilde{\lambda}}{|\data|} \|\theta\|^2,
% \]
% where $Z$ is a set of i.i.d samples from the index distribution $p_z$ (\mohammad{We haven't defined $p_z$. Probably just saying "from the space of one-hot vectors."}.

% We sweep over ensemble size $\in \{1, 3, 10, 30, 100\}$, $L_2$ weight decay scale $ \lambda \in \{10^{-4}, 10^{-3}, 10^{-2}, 0.1, 1, 10, 100\}$ and \textit{adaptive} weight decay $\tilde{\lambda} \in \{ \lambda, \sqrt{\rho} d \lambda \}$ where $d$ is the data dimension and $\rho$ is the data temperature.
%We find that the best value of $\lambda=1$ with adaptive weight decay.

%%%%%%%%%%%%%%%%%%%%%%%%%%%%%%%%%%%%%%%%%%%%%%%%%%%%%%%%%%%%%%%%% MC dropout
\subsection{Dropout}

We follow \cite{Gal2016Dropout} to build a \texttt{droput} agent for posterior approximation. The agent applies dropout on each layer of a fully connected MLP with ReLU activation and optimizes the network using the cross-entropy loss combined with $L_2$ weight decay. The $L_2$ weight decay scale is chosen to be either  $\frac{l^2}{2T}(1 - p_{\textrm{drop}})$ or $\frac{d \sqrt{\beta} l}{T}$ where $p_{\textrm{drop}}$ is the dropping probability, $d$ is the input dimension, $\beta$ is the temperature of the data generating process, and $T$ is the size of the training dataset. We sweep over dropout rate $p_{\textrm{drop}} \in \{0, 0.1, 0.2, 0.3, 0.4, 0.5, 0.6, 0.7, 0.8, 0.9\}$, length scale (used for $L_2$ weight decay) $l \in \{1, 3, 10\}$, number of neural network layers $\in \{2, 3\}$, and hidden layer size $\in \{50, 100\}$.
The implementation and hyperparameter sweeps for the \texttt{dropout} agent can be found in our open source code under the path \url{/agents/factories/dropout.py}. 

% \[
% \Lc (\theta, \tilde{\data}) = - \frac{1}{ |\tilde{\data}|} \sum_{(x, y) \in \tilde{\data}} \frac{1}{|Z|}\sum_{z \in Z} \mathrm{XEntropy} \left( y, f_\theta(x, z) \right) +
% \frac{\tilde{\lambda}}{|\data|} \|\theta\|^2,
% \]
% where $Z$ is a set of i.i.d samples from the space of Bernoulli masks for the network layers.

%%%%%%%%%%%%%%%%%%%%%%%%%%%%%%%%%%%%%%%%%%%%%%%%%%%%%%%%%%%%%%%%% BBB
\subsection{Bayes-by-backprop}

We follow \cite{blundell2015weight} to build a \texttt{bbb} agent for posterior approximation. We consider a scale mixture of two zero-mean Gaussian densities as the prior. The Gaussian densities have standard deviations $\sigma_1$ and $\sigma_2$, and they are mixed with probabilities $p$ and $1 - p$, respectively. We sweep over $\sigma_1 \in \{0.3, 0.5, 0.7, 1, 2, 4\}$, $\sigma_2 \in \{0.3, 0.5, 0.7\}$, $p \in \{0, 0.5, 1\}$, learning rate $\in \{10^{-3}, 3 \times 10^{-3}\}$, number of training steps $\in \{1000, 2000\}$,  number of neural network layers $\in \{2, 3\}$, hidden layer size $\in \{50, 100\}$, and the ratio of the complexity cost to the likelihood cost $\in \{1, d\sqrt{\beta}\}$, where $d$ is the input dimension and $\beta$ is the temperature of the data generating process. The implementation and hyperparameter sweeps for the \texttt{bbb} agent can be found in our open source code under the path \url{/agents/factories/bbb.py}.

%%%%%%%%%%%%%%%%%%%%%%%%%%%%%%%%%%%%%%%%%%%%%%%%%%%%%%%%%%%%%%%%% SGMCMC
\subsection{Stochastic gradient Langevin dynamics}
\label{sec:agents-sgmcmc}

We follow \cite{welling2011bayesian} to implement a \texttt{sgmcmc} agent using stochastic gradient Langevin dynamics (SGLD). We consider two versions of SGLD, one with momentum and other without the momentum. We consider independent Gaussian prior on the neural network parameters where the prior variance is set to be
\begin{equation*}
    \sigma^2 = \lambda\frac{T}{d \sqrt{\beta}},
\end{equation*}
where $\lambda$ is a hyperparameter that is swept over $\{0.0025, 0.01, 0.04\}$, $d$ is the input dimension, $\beta$ is the temperature of the data generating process, and $T$ is the size of the training dataset. We consider a constant learning rate that is swept over $\{10^{-4}, 5 \times 10^{-4}, 10^{-3}, 5 \times 10^{-3}\}$. For SGLD with momentum, the momentum decay term is always set to be $0.9$.  The number of training batches is $5 \times 10^5$ with burn-in time of $10^5$ training batches. We save a model every 1000 steps after the burn-in time and use these models as an ensemble during the evaluation.
The implementation and hyperparameter sweeps for the \texttt{sgmcmc} agent can be found in our open source code under the path \url{/agents/factories/sgmcmc.py}. 

%%%%%%%%%%%%%%%%%%%%%%%%%%%%%%%%%%%%%%%%%%%%%%%%%%%%%%%%%%%%%%%%% Randomized prior
\subsection{Ensemble+}
\label{sec:agents-ensemble_rpf}

We implement the \texttt{ensemble+} agent using deep ensembles with randomized prior functions \citep{osband2018rpf} and bootstrap sampling \citep{osband2015bootstrapped}.
Similar to the vanilla ensemble agent in Section \ref{sec:agents-vanila_ensemble}, we consider $L_2$ weight scale to be either $\lambda \frac{1}{MT}$ or $\lambda \frac{d\sqrt{\beta}}{M T}$.
We sweep over ensemble size $M \in \{1, 3, 10, 30, 100\}$ and $\lambda \in \{0.1, 0.3, 1, 3, 10\}$.
The randomized prior functions are sampled exactly from the data generating process, and we use a  prior scale of $3/\sqrt{\beta}$. In addition, we sweep over bootstrap type ({\rm none}, {\rm exponential}, {\rm bernoulli}).

Note that an ensemble+ agent is obtained by an addition of a prior network to the ensemble agent. We find that the addition of randomized prior functions is crucial for improvement in performance over vanilla deep ensembles in terms of the quality of joint predictions. The implementation and hyperparameter sweeps for the \texttt{ensemble+} agent can be found in our open source code under the path \url{/agents/factories/ensemble_plus.py}. 

% \textcolor{red}{CITE \cite{he2020bayesian} AND EXPLAIN CONNECTION TO THE DIFFERENCE BETWEEN ENSEMBLE AND ENSEMBLE+, AND ALSO MENTION THE CONNECTION BETWEEN THAT WORK AND THE PRIOR NETWORKS PAPER}

%%%%%%%%%%%%%%%%%%%%%%%%%%%%%%%%%%%%%%%%%%%%%%%%%%%%%%%%%%%%%%%%% Hypermodel
\subsection{Hypermodel}

We follow \cite{Dwaracherla2020Hypermodels} to build a \texttt{hypermodel} agent for posterior approximation. We consider a linear hypermodel over a 2-layer MLP base model. We sweep over index dimension $\in \{1, 3, 5, 7\}$. The $L_2$ weight decay is chosen to be either $\lambda\frac{1}{T}$ or $\lambda \frac{d \sqrt{\beta} }{T}$ with $\lambda \in \{0.1, 0.3, 1, 3, 10\}$, where $d$ is the input dimension, $\beta$ is the temperature of the data generating process, and $T$ is the size of the training dataset. We sweep over bootstrap type ({\rm none}, {\rm exponential}, {\rm bernoulli}). We use an additive prior which is a linear hypermodel prior over an MLP base model, which is similar to the generating process, with number of hidden layers in $\{1, 2\}$, $10$ hidden units in each layer, and prior scale from $\{1/\sqrt{\beta}, 1/\beta\}$.
The implementation and hyperparameter sweeps for the \texttt{hypermodel} agent can be found in our open source code under the path \url{/agents/factories/hypermodel.py}.

%%%%%%%%%%%%%%%%%%%%%%%%%%%%%%%%%%%%%%%%%%%%%%%%%%%%%%%%%%%%%%%%% Non-Parametric
\subsection{Non-parametric classifiers}
\label{sec:non-parametric}
K-nearest neighbors (k-NN) \citep{cover1967nearest} and random forest classifiers \citep{friedman2017elements} are simple and cheap off-the-shelf non-parametric baselines \citep{murphy2012machine, scikit-learn}. The `uncertainty' in these classifiers arises merely from the fact that they produce distributions over the labels and as such we do not expect them to perform well relative to more principled approaches. Moreover, these methods have no capacity to model $\KL^\tau$ for $\tau > 1$. For the \texttt{knn} agent we swept over the number of neighbors $k \in \{1, 5, 10, 30, 50, 100\}$ and the weighting of the contribution of each neighbor as either uniform or based on distance. For the \texttt{random\_forest} agent  we swept over the number of trees in the forest $\{10, 100, 1000\}$, and the splitting criterion which was either the Gini impurity coefficient or the information gain. To prevent infinite values in the KL we truncate the probabilities produced by these classifiers to be in the interval $[0.01, 0.99]$.
The implementation and hyperparameter sweeps for the \texttt{knn} and \texttt{random\_forest} agents can be found in our open source code under the paths \url{/agents/factories/knn.py} and \url{/agents/factories/random_forest.py}.

% %%%%%%%%%%%%%%%%%%%%%%%%%%%%%%%%%%%%%%%%%%%%%%%%%%%%%%%%%%%%%%%%% Deep-kernel
\subsection{Gaussian process with learned kernel}
\label{sec:deep-kernel}
A neural network takes input $X_t \in \mathcal{X}$ and produces output $Z_{t+1} = W\phi_\theta(X_t) + b\in \reals^{K}$, where $W \in \reals^{K \times m}$ is a matrix, $b \in \Real^{K}$ is a bias vector, and $\phi_\theta: \mathcal{X} \rightarrow \Real^{m}$ is the output of the penultimate layer of the neural network. In the case of classification the output $Z_{t+1}$ corresponds to the logits over the class labels, i.e., $\hat Y_{t+1} \propto \exp(Z_{t+1})$. The neural network should learn a function that maps the input into a space where the classes are linearly distinguishable. In other words, the mapping that the neural network is learning can be considered a form of \emph{kernel} \citep{scholkopf2002learning}, where the kernel function $k: \mathcal{X} \times \mathcal{X} \rightarrow \reals$ is simply $k(X, X^\prime) = \phi_\theta(X)^\top \phi_\theta(X^\prime)$. With this in mind, we can take a \emph{trained} neural network and consider the learned mapping to be the kernel in a Gaussian process (GP) \citep{rasmussen2003gaussian}, from which we can obtain approximate uncertainty estimates. 
%Without loss of generality we assume that the output of the penultimate layer is normalized, i.e., $\|\phi_\theta(x)\|_2 = 1$ for any input $x$.
%\[
%y^{(i)} = w_i^\top \phi_\theta(x) + b^{(i)} + \epsilon_i
%\]
Concretely, let $\Phi_{0:T-1} \in \reals^{T \times m}$ be the matrix corresponding to the $\phi_\theta(X_t)$, $t=0, \ldots, T-1$, vectors stacked row-wise and let $\Phi_{T:T+\tau-1}\in \reals^{\tau \times m}$ denote the same quantity for the test set. We can write the kernel function evaluated on the training and test datasets using these matrices. Fix index $i \in \{0, \ldots, K-1\}$ to be a particular class index. A GP models the joint distribution over the dataset to be a multi-variate Gaussian, i.e.,
\[
 \begin{bmatrix}
 Z^{(i)}_{1:T} \\
 Z^{(i)}_{T+1:T+\tau}
 \end{bmatrix}
 \sim
 \mathcal{N} \left(
 \begin{bmatrix}
 \mu^{(i)}_{1:T} \\
 \mu^{(i)}_{T+1:T+\tau}
 \end{bmatrix},
 \begin{bmatrix}
 \sigma^2 I + \Phi_{0:T-1} \Phi_{0:T-1}^\top & \Phi_{T:T+\tau-1} \Phi_{0:T-1}^\top \\
 \Phi_{0:T-1} \Phi_{T:T+\tau-1}^\top & \Phi_{T:T+\tau-1} \Phi_{T:T+\tau-1}^\top
 \end{bmatrix}
 \right)
 \]
 where $\sigma \ > 0$ models the noise in the training data measurement and $\mu^{(i)}_{1:T}$, $\mu^{(i)}_{T+1:T+\tau}$ are the means under the GP. The conditional distribution is given by
 \begin{align*}
 P(Z^{(i)}_{T+1:T+\tau} \mid Z^{(i)}_{1:T}, X_{0:T+\tau-1})
 &= \mathcal{N} \left(\mu_{T+1:T+\tau \mid 1:T}^{(i)}, \Sigma_{T+1:T+\tau \mid 1:T} \right)
 \end{align*}
 where 
 \begin{align*}
     \Sigma_{T+1:T+\tau \mid 1:T} &= \Phi_{T:T+\tau-1} \Phi_{T:T+\tau-1}^\top - \Phi_{T:T+\tau-1} \Phi_{0:T-1}^\top (\sigma^2 I + \Phi_{0:T-1} \Phi_{0:T-1}^\top)^{-1} \Phi_{0:T-1} \Phi_{T:T+\tau-1}^\top.
 \end{align*}
and rather than use the GP to compute $\mu_{T+1:T+\tau \mid 0:T}^{(i)}$ (which would not be possible since we do not observe the true logits) we just take it to be the output of the neural network when evaluated on the test dataset. 
%, that is,
%\[
%\mu_{T+1:T+\tau \mid 1:T} := \phi_\theta(X_{T : T+\tau - 1}) W^\top + b %\mathbf{1}^\top \in \reals^{\tau \times m}.
%\]
The matrix being inverted in the expression for $\Sigma_{T+1:T+\tau \mid 0:T}$ has dimension $T \times T$, which may be quite large. We use the Sherman-Morrison-Woodbury identity to rewrite it as follows \citep{woodbury1950inverting}
 \begin{align*}
     \Sigma_{T+1:T+\tau \mid 0:T}
     &= \Phi_{T:T+\tau-1} (I -  \Phi_{0:T-1}^\top (\sigma^2 I + \Phi_{0:T-1} \Phi_{0:T-1}^\top)^{-1} \Phi_{0:T-1} )\Phi_{T:T+\tau-1}^\top \\
     &= \sigma^2 \Phi_{T:T+\tau-1} (\sigma^2 I + \Phi_{0:T-1}^\top \Phi_{0:T-1} )^{-1}\Phi_{T:T+\tau-1}^\top,
 \end{align*}
which instead involves the inverse of an $m \times m$ matrix, which may be much smaller. If we perform a Cholesky factorization of positive definite matrix $(\sigma^2 I + \Phi_{0:T-1}^\top \Phi_{0:T-1}) = L L^\top$ then the samples for all logits simultaneously can be drawn by first sampling $\zeta \in \reals^{m \times K}$, with each entry drawn IID from $\mathcal{N}(0, 1)$,
then forming
 \[
 \hat Y_{T+1:T+\tau} \propto \exp(\mu_{T+1:T+\tau \mid 1:T} + \sigma \Phi_{T:T+\tau-1} L^{-\top} \zeta).
 \]
The implementation and hyperparameter sweeps for the \texttt{deep\_kernel} agent can be found in our open source code under the path \url{/agents/factories/deep_kernel.py}.

%%%%%%%%%%%%%%%%%%%%%%%%%%%%%%%%%%%%%%%%%%%%%%%%%%%%%%%%%%%%%%%%% Other agents
\subsection{Other agents}
\label{sec:other_agents}

In our paper we have made a concerted effort to include representative and canonical agents across different families of Bayesian deep learning and adjacent research.
In addition to these implementations, we performed extensive tuning to make sure that each agent was given a fair shot.
However, with the proliferation of research in this area, it was not possible for us to evaluate all competiting approaches.
We hope that, by opensourcing the Neural Testbed, we can allow researchers in the field to easily assess and compare their agents to these baselines.

For example, we highlight a few recent pieces of research that might be interesting to evaluate in our setting.
Of course, there are many more methods to compare and benchmark.
We leave this open as an exciting area for future research.
\begin{itemize}[leftmargin=*]
    \item \textbf{Neural Tangent Kernel Prior Functions} \citep{he2020bayesian}.
    Proposes a specific type of prior function in \textit{ensemble+} inspired by connections to the neural tangent kernel.
    
    \item \textbf{Functional Variational Bayesian Neural Networks} \citep{sun2019functional}. 
    Applies variational inference directly to the function outputs, rather than weights like \texttt{bbb}.
    
    \item \textbf{Variational normalizing flows} \citep{rezende2015variational}.
    Applies variational inference over a more expressive family than \texttt{bbb}.
    
    \item \textbf{No U-Turn Sampler} \citep{hoffman2014no}.
    Another approach to \texttt{sgmcmc} that attempts to compute the posterior directly, computational costs can grow large.
\end{itemize}

%%%%%%%%%%%%%%%%%%%%%%%%%%%%%%%%%%%%%%%%%%%%%%%%%%%%%%%%%%%%%%%%% TESTBED
%%%%%%%%%%%%%%%%%%%%%%%%%%%%%%%%%%%%%%%%%%%%%%%%%%%%%%%%%%%%%%%%%
\section{Testbed results}
\label{app:testbed-results}
In this section, we provide the complete results of the performance of benchmark agents on the Testbed, broken down by the temperature setting, which controls the SNR, and the size of the training dataset.
We select the best performing agent, based on aggregated score $\KL^1 + \KL^{10}/10$, within each agent family and plot $\KL^1$ and $\KL^{10}$ with the performance of an MLP agent as a reference. We also provide a plot comparing the training time of different agents.

\subsection{Visualizing \texttt{ensemble} vs \texttt{ensemble+}}
\label{app:testbed-visualize}

\begin{figure}[!ht]
    \centering
    \begin{subfigure}[b]{0.18\textwidth}
        \centering
        \includegraphics[height=1.2in]{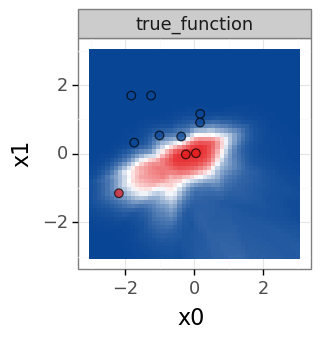}
        \caption{True model.}
    \end{subfigure}%
    ~ 
    \begin{subfigure}[b]{0.8\textwidth}
        \centering
        \includegraphics[height=1.2in]{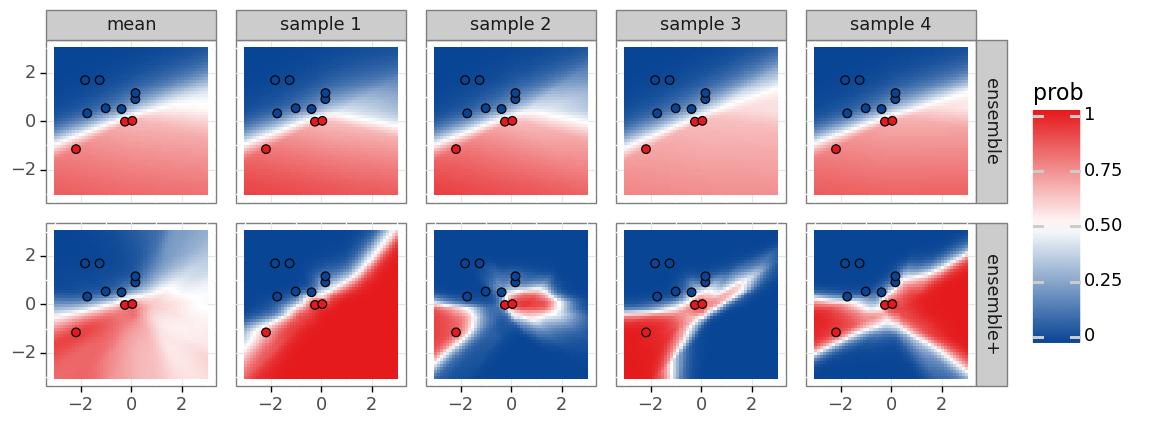}
        \caption{Agent samples: only ensemble+ produces diverse decision boundaries.}
    \end{subfigure}
    \caption{Visualization of the predictions of ensemble and ensemble+ agents.}
    \label{fig:synthetic_ensemble_plot_enn}
\end{figure}

% You get to visualize the problem
Figure~\ref{fig:synthetic_ensemble_plot_enn} provides additional intuition into \textit{how} the randomized prior functions are able to drive improved performance.
Figure~\ref{fig:synthetic_ensemble_plot_enn}a shows a sampled generative model from our Testbed, with the training data shown in red and blue circles.
Figure~\ref{fig:synthetic_ensemble_plot_enn}b shows the mean predictions and $4$ randomly sampled ensemble members from each agent (top=\texttt{ensemble}, bottom=\texttt{ensemble+}).
We see that, although the agents mostly agree in their mean predictions, \texttt{ensemble+} produces more diverse sampled outcomes enabled by the addition of randomized prior functions.
In contrast, \texttt{ensemble} produces similar samples, which may explain why its performance is close to baseline \texttt{mlp} in this setting.

\subsection{Performance breakdown}

\begin{figure}[!ht]
    \centering
    \includegraphics[width=.99\textwidth]{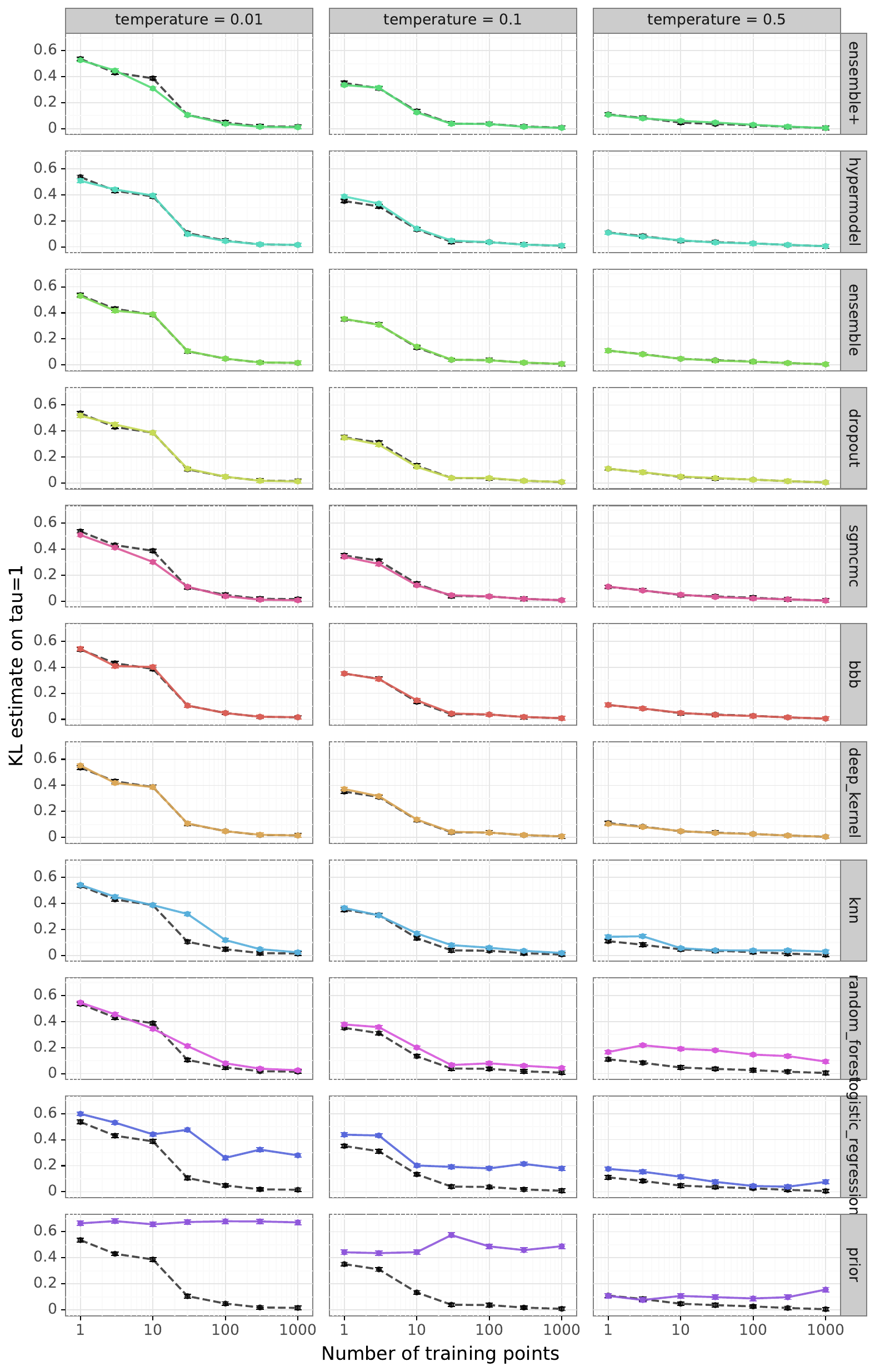}
    \caption{Performance of benchmark agents on the Testbed evaluated on $\tau=1$, compared against the MLP baseline.}
    \label{fig:testbed_breakdown_tau_1}
\end{figure}

\begin{figure}[!ht]
    \centering
    \includegraphics[width=.99\textwidth]{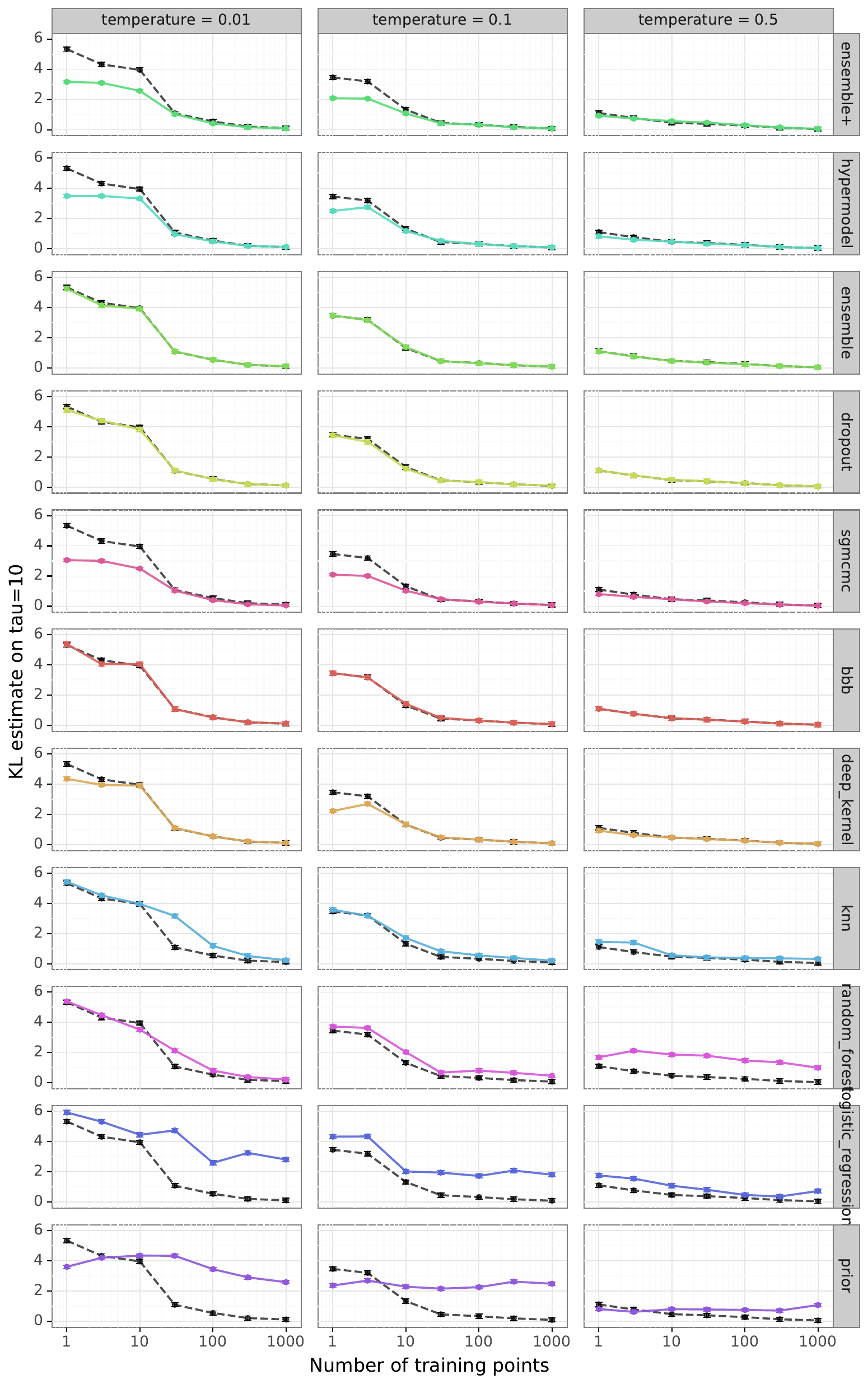}
    \caption{Performance of benchmark agents on the Testbed evaluated on $\tau=10$, compared against the MLP baseline.}
    \label{fig:testbed_breakdown_tau_100}
\end{figure}

Figures~\ref{fig:testbed_breakdown_tau_1} and \ref{fig:testbed_breakdown_tau_100} show the KL estimates evaluated on $\tau = 1$ and $\tau = 10$, respectively.
For each agent, for each SNR regime, for each number of training points we plot the average KL estimate from the Testbed.
In each plot, we include the ``baseline'' \texttt{mlp} agent as a black dashed line to allow for easy comparison across agents.
A detailed description of these benchmark agents can be found in Appendix~\ref{app:agents}.

\subsection{Training time}
\label{app:testbed-speed}

\begin{figure}[!ht]
    \centering
    \includegraphics[width=0.6\textwidth]{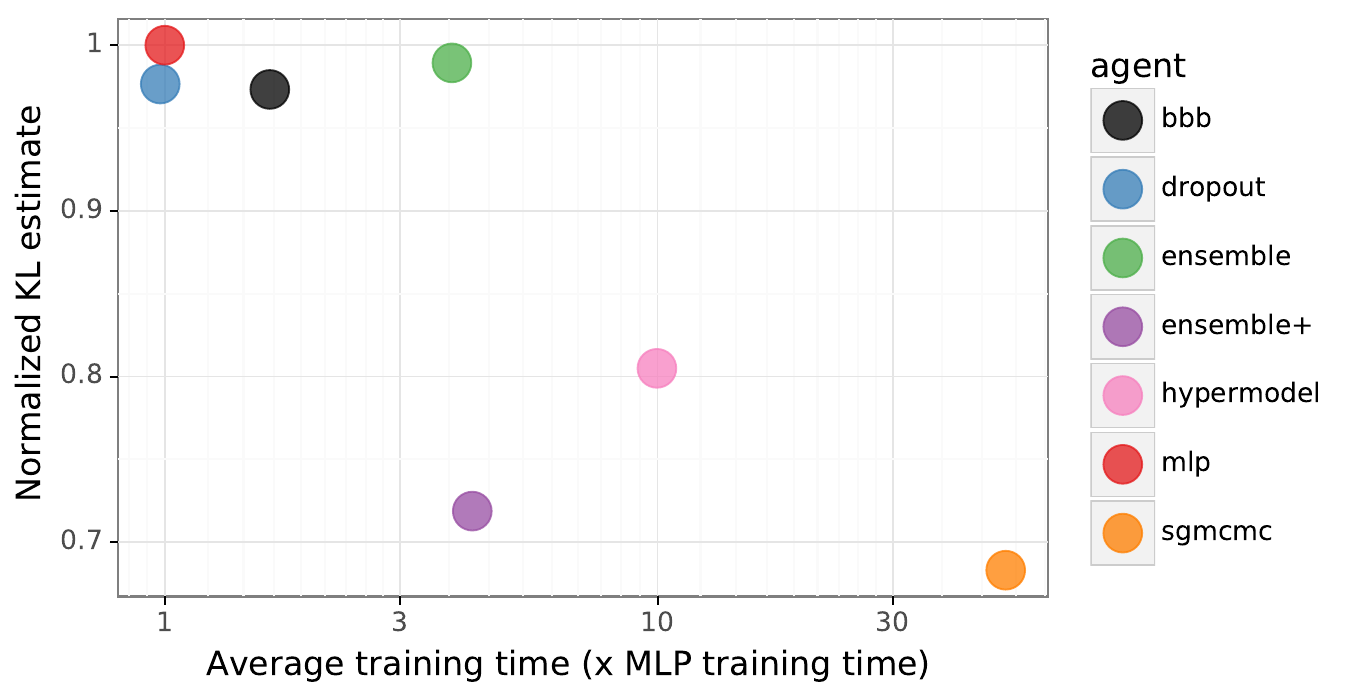}
    \caption{Normalized $\KL^{10}$ vs training time of different agents}
    \label{fig:training_time}
\end{figure}

Figure \ref{fig:training_time} shows a plot comparing the $\KL^{10}$ and training time of different agents normalized with that of an MLP. The parameters of each agent are selected to maximize the $\KL^{10}$. We can see that \texttt{sgmcmc} agent has the best performance, but at the cost of more training time (computation). Both \texttt{ensemble+} and \texttt{hypermodel} agents have similar performance as \texttt{sgmcmc} with lower training time. We trained our agents on CPU only systems.

%%%%%%%%%%%%%%%%%%%%%%%%%%%%%%%%%%%%%%%%%%%%%%%%%%%%%%%%%%%%%%%%% BANDIT
%%%%%%%%%%%%%%%%%%%%%%%%%%%%%%%%%%%%%%%%%%%%%%%%%%%%%%%%%%%%%%%%%
\newpage

\section{Sequential Decision Problems}
\label{app:sequential}

This section provides supplementary information for the sequential decision problems in Section \ref{sec:sequential}.
All of the code necessary to reproduce the experiments is opensourced in the \url{/bandit/} directory.

%%%%%%%%%%%%%%%%%%%%%%%%%%%%%%%%%%%%%%%%%%%%%%%%%%%%%%%%%%%%%%% Problem
\subsection{Problem formulation}
\label{app:sequential_problem}

We consider bandit problems derived from the testbed and evaluate the agents using Algorithm \ref{alg:bandit-evaluation} for which we need to specify prior on the environment $\Prob(\environment \in \cdot)$, input distribution $P_X$, and the number of actions $N$. We choose input distribution $P_X = \mathcal{N}(0, I_d)$, where $d$ is the input dimension. We sweep over $d \in \{2, 10, 50\}$ and choose the number of actions to be $N = 20\, d$, i.e.,  for input dimensions $\{2, 10, 50\}$ we have $\{40, 200, 1000\}$ actions respectively. We use the same prior distribution of environments as in Appendix \ref{app:testbed_pseudo_code} with a fixed temperature of $0.1$. For each setting, we run for $50,000$ time steps ($T=50,000$) and with $20$ random seeds ($J=20$).

\begin{algorithm}[!ht]
\caption{Evaluation on Bandit Problem}
\label{alg:bandit-evaluation}
% \setstretch{1.1}
\begin{algorithmic}
\REQUIRE Evaluation on bandit problem requires the following inputs
\begin{enumerate}
\item Distribution over the environment $\Prob(\environment \in \cdot)$, input distribution $P_X$, and the number of actions $N$. 
% \item agent with belief distribution over $f_\theta$
\item Agent $f_\theta$
\item Number of time steps $T$
\item Number of sampled problems $J$
% and \ref{alg:rp-estimation}
\end{enumerate}
\vspace{0.3em}
\FOR{$j=1, 2, \ldots, J$}
\vspace{0.3em}
\STATE \textbf{Step 1: Sample environment and action set}
    \STATE \hspace{1em} 1. Sample environment $\environment \sim \Prob(\environment \in \cdot)$
    \STATE \hspace{1em} 2. Sample a set $\mathcal{X}$ of $N$ actions $x_1, x_2, \ldots, x_{N}$ i.i.d. from $P_X$ 
    \STATE \hspace{1em} 3. Obtain the mean rewards corresponding to actions in $\mathcal{X}$ conditioned on the environment
    \[
    \overline{R}_x = \Prob(Y_{t+1} = 1| \environment, X_t = x), \quad \forall x \in \mathcal{X}
    \]
    \STATE \hspace{1em} 4. Compute the optimal expected reward $\overline{R}_* = \max_{ x \in \mathcal{X} } \overline{R}_x$
    % \STATE \hspace{1em} 4. choose the training dataset as
    % $
    % \data_T = \left \{ \left(X_t, Y_{t+1} \right), \, t=0, \ldots, T-1 \right \}$
\vspace{0.3em}
\STATE \textbf{Step 2: Agent interaction with the environment}
    \STATE \hspace{1em} Initialize the data buffer $\data_0 = \{\}$
    \STATE \hspace{1em} \textbf{for} $t=1,2,\ldots, T$ \textbf{do}
        \STATE \hspace{2em} 1. Update agent $f_{\theta_t}$ belief distribution based on the data in the buffer $\data_{t-1}$
        \STATE \hspace{2em} 2. TS action selection scheme:
            % \STATE \hspace{3em} i. sample $f_{\theta_t}$ from agent belief distribution
            \STATE \hspace{3em} i. Sample $\hat{\environment}_t$ from the agent belief distribution 
            \[ \hat{\environment}_t \sim \Prob\left( \hat{\environment} \in \cdot | {\theta_t}\right) \]
            \STATE \hspace{3em} ii. Act greedily based on $\hat\environment_t$
            \[
            % A_t \in \arg\max_{a \in \Ac} f_{\theta_t}(a)
            X_t \in \arg\max_{x \in \mathcal{X}} \Prob(\hat Y_{t+1} =1 | \hat\environment_t, X_t=x )
            \]
            \STATE \hspace{3em} iii. Generate observation $Y_{t+1}$ based on action $X_t$
            \[ 
            Y_{t+1} \sim \Prob\left(Y_{t+1} \in \cdot | \environment, X_t = X_t \right)
            \]
        \STATE \hspace{2em} 3. Update the buffer $\data_t = \data_0 \cup {(X_t, Y_{t+1})}$
\vspace{0.3em}
        \STATE \hspace{1em} \textbf{end for}
    \STATE \hspace{1em} Compute the total regret incurred in $T$ time steps 
    \[
    {\rm Regret}_j(T) = \sum_{t=1}^T \left(\overline{R}_* - \overline{R}_{X_t} \right) 
    \]  
    \vspace{0.3em}
\ENDFOR \\
\RETURN $\frac{1}{J} \sum_{j=1}^J {\rm Regret}_j(T)$
\end{algorithmic}
% \setstretch{1}
\end{algorithm}

%%%%%%%%%%%%%%%%%%%%%%%%%%%%%%%%%%%%%%%%%%%%%%%%%%%%%%%%%%%%%%% Agents
\subsection{Agent definition}
\label{app:sequential_agent}

In Appendix \ref{app:agents}, we described benchmark agents in our testbed. Among these agents, we use {\tt mlp}, {\tt ensemble}, {\tt dropout}, {\tt bbb},  {\tt ensemble+}, and {\tt hypermodel} agents for sequential decision problems. For all the agents we use the hyper parameters specified by default, in the source code, at the path \url{/agents/factories/}.
The default hyperparameters of each agent correspond to be the ones that minimize the aggregated KL score $\KL^{\rm agg} = \KL^1 + \KL^{10}/10$. As the agent interacts with the environment, the amount of data the agent has observed keeps growing. Due to this we tune the regularization term based on the number of time steps agent has interacted with the environment.
For {\tt mlp}, {\tt ensemble}, {\tt ensemble+}, and {\tt hypermodel} agents we use an $L_2$ weight decay of $\lambda\frac{2 \sqrt{\beta}}{t}$, where $\beta$ is the temperature, $t$ is the number of the time steps the agent has interacted with the environment, and $\lambda$ is the default weight scale of the agent. For {\tt dropout} we choose the $L_2$ weight decay as $\frac{2 \sqrt{\beta}l}{t}$, where $l$ is the default length scale used in the dropout agent. For {\tt bbb} we scale the prior term by $\frac{1}{t}$.
As described above, all hyperparmeters are chosen to be the ones which minimize the aggregated  KL score $\KL^{\rm agg} = \KL^1 + \frac{1}{10}\KL^{10}$.

%%%%%%%%%%%%%%%%%%%%%%%%%%%%%%%%%%%%%%%%%%%%%%%%%%%%%%%%%%%%%%% Results

\subsection{Results}
\label{app:sequential_results}

\begin{figure}[!ht]
    \centering
    \includegraphics[width=0.9\textwidth]{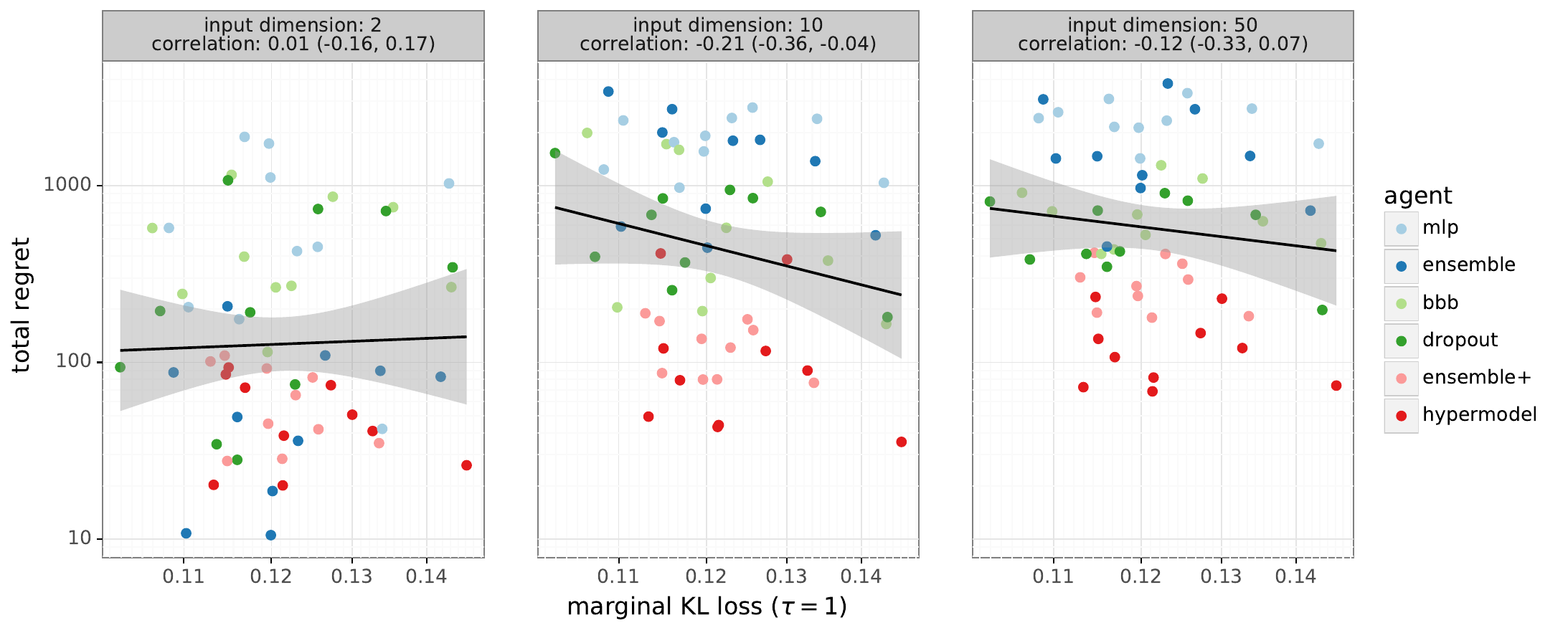}
    \caption{Testbed marginal performance $\KL^1$ is not significantly positively correlated with sequential decision performance. This result is robust across input dimensions 2, 10, and 50.}
    \label{fig:bandit_corr_input_dim_tau_1}
\end{figure}

\begin{figure}[!ht]
    \centering
    \includegraphics[width=0.9\textwidth]{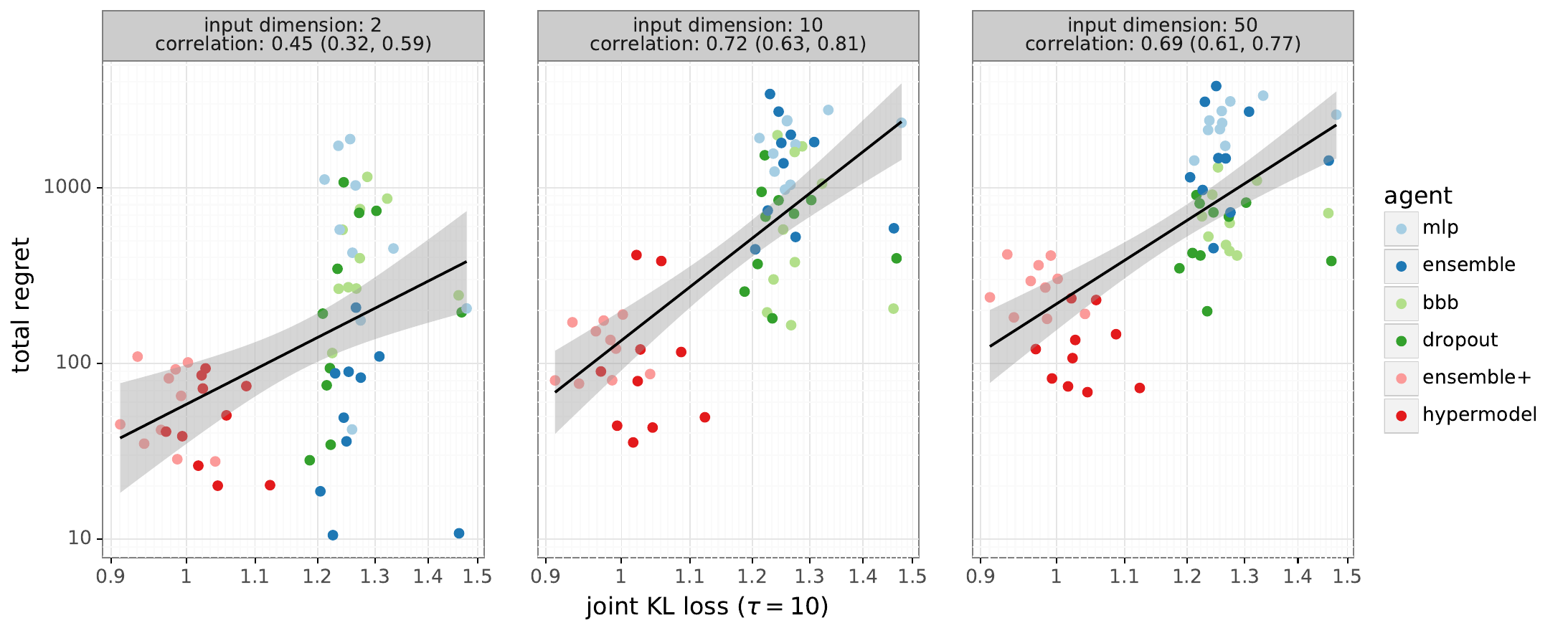}
    \caption{\centering Testbed joint performance $\KL^{10}$ is significantly positively correlated with sequential decision performance. This result is robust across input dimensions 2, 10, and 50.}
    \label{fig:bandit_corr_input_dim_tau_10}
\end{figure}

Figures \ref{fig:bandit_tau_1} and \ref{fig:bandit_tau_10} shows the correlation between performance on testbed performance and sequential decision problems with an input dimension of $50$. Different points of an agent in these figures corresponds to different random seeds for the testbed and sequential problems. We can see that performance on sequential decision problems is strongly correlated with testbed joint performance $\tau=10$ and not correlated with the testbed marginal performance. In Figures \ref{fig:bandit_corr_input_dim_tau_1} and \ref{fig:bandit_corr_input_dim_tau_10} we show a similar correlation plots between testbed performance and sequential decision problems across different input dimensions for sequential decision problems. We can see that performance on sequential decision problems has clear correlation with testbed joint performance $\tau=10$, and no correlation with testbed marginal performance $\tau=1$, across all the input dimensions considered.

These results offer empirical evidence that practical deep learning approaches separated by the quality of their joint predictions, but not their marginals, can lead to differing performance in downstream tasks. In addition, we show that our simple 2D testbed can provide insights that scale to much higher dimension problems.

\end{document}